%% file: main.tex
\documentclass[10pt,twocolumn,letterpaper]{article}
\usepackage[accsupp]{axessibility} 
\PassOptionsToPackage{table,dvipsnames}{xcolor}
\usepackage[pagenumbers]{cvpr} 

\definecolor{cvprblue}{rgb}{0.21,0.49,0.74}
\usepackage[pagebackref=false,breaklinks,colorlinks,allcolors=cvprblue]{hyperref}

\input{preamble}
\usepackage{times}  
\usepackage{helvet}  
\usepackage{courier}  
\usepackage[hyphens]{url}  
\usepackage{graphicx} 
\usepackage{natbib}  
\usepackage{caption} 
\usepackage{algorithm}
\usepackage{algorithmic}
\usepackage{comment}
\usepackage{graphicx}
\usepackage{amsmath}
\usepackage{amssymb}
\usepackage{booktabs}
\usepackage{enumitem}
\usepackage{makecell}
\usepackage{url}
\usepackage{tablefootnote}

\usepackage[accsupp]{axessibility} 
\usepackage{booktabs, multirow} 

\definecolor{lora_orange}{RGB}{233, 113, 50}
\definecolor{lora_blue}{RGB}{78, 149, 217}

\definecolor{lightgray}{gray}{0.9}
\definecolor{linecolor}{RGB}{204, 204, 255}

\usepackage[capitalize]{cleveref}
\crefname{section}{Sec.}{Secs.}
\Crefname{section}{Section}{Sections}
\Crefname{table}{Table}{Tables}
\crefname{table}{Tab.}{Tabs.}
\definecolor{mygray}{gray}{.9}
\definecolor{mygray1}{gray}{.92}
\definecolor{evaunit01green}{RGB}{82,208,83}
\definecolor{lowred}{RGB}{238,18,137}

\definecolor{lowerred}{RGB}{255,110,180}

\newcommand{\dplus}[1]{\fontsize{6pt}{0.1em}\selectfont (\textbf{\textcolor{lowred}{#1}})}

\definecolor{defaultcolor}{RGB}{12,127,17}

\def\recon{{\scshape ReCon}}

\def\paperID{2355} 
\def\confName{CVPR}
\def\confYear{2025}

\title{\textbf{\texttt{\textcolor{lora_blue}{Point}\textcolor{
lora_orange}{LoRA}}}: Low-Rank Adaptation with Token Selection \\ for Point Cloud Learning}

\author{Song Wang$^{1,2}$, \ \ \  Xiaolu Liu$^1$, \ \ \ Lingdong Kong$^2$, \ \ \  Jianyun Xu$^3\thanks{Project leader.}$,  \ \ \ Chunyong Hu$^3$, \\ Gongfan Fang$^2$, \ \ \ Wentong Li$^4$, \ \ \ Jianke Zhu$^{1}\footnotemark[2]$, \ \ \ Xinchao Wang$^{2}\thanks{Corresponding authors.}$\\
	$^1$ZJU \ \ \
	$^2$NUS \ \ \
    $^3$AD Lab, CaiNiao, Alibaba \ \ \
    $^4$NUAA \ \ \
    \\ 
	{\tt\small \{songw, jkzhu\}@zju.edu.cn, xinchao@nus.edu.sg}
}

\begin{document}
\maketitle

\input{sections/0_abstract}
\input{sections/1_intro}

\input{sections/2_related_work}

\input{sections/3_prelim}
\input{sections/4_method}
\input{sections/5_exps}

\input{sections/6_conclusion}

\clearpage\clearpage

\section*{Acknowledgments}
This work is supported in part by the National Natural Science Foundation of China under Grant No. 62376244, in part by the Zhejiang Provincial Natural Science Foundation of China under Grant No. LD24F030001,
by the Information Technology Center and State Key Lab of CAD\&CG, Zhejiang University,
and by the National Research Foundation, Singapore, and Cyber Security Agency of Singapore under its National Cybersecurity R\&D Programme and CyberSG R\&D Cyber Research Programme Office (Award: CRPO-GC1-NTU-002).
Lingdong Kong is supported by the Apple Scholars in AI/ML Ph.D. Fellowship program.

{
    \small
    \bibliographystyle{ieeenat_fullname}
    \bibliography{main}
}

\input{supp}

\end{document}

%% file: preamble.tex
\usepackage{graphicx}
\usepackage{amsmath}
\usepackage{amssymb}
\usepackage{booktabs}
\usepackage{multirow}
\usepackage{mathrsfs}
\usepackage{appendix}
\usepackage{subcaption}
\usepackage{pifont}

\definecolor{evaunit01green}{RGB}{82,208,83}
\definecolor{rowunit}{RGB}{128,128,255}
\definecolor{evaunit01green}{RGB}{54,125,189}
\newcommand{\evagreen}[1]{\textcolor{evaunit01green}{#1}}
\newcommand{\dtplus}[1]{\fontsize{6pt}{0.1em}\selectfont (\textbf{\evagreen{#1}})}

\usepackage{tabularx}
\usepackage[capitalize]{cleveref}
\crefname{section}{Sec.}{Secs.}
\Crefname{section}{Section}{Sections}
\Crefname{table}{Table}{Tables}
\crefname{table}{Tab.}{Tabs.}

\newcommand{\yxq}[1]{{\color{green} }}

\def\recon{{\scshape ReCon}}

%% file: sections/0_abstract.tex
\begin{abstract}
Self-supervised representation learning for point cloud has demonstrated effectiveness in improving pre-trained model performance across diverse tasks. However, as pre-trained models grow in complexity, fully fine-tuning them for downstream applications demands substantial computational and storage resources. Parameter-efficient fine-tuning (PEFT) methods offer a promising solution to mitigate these resource requirements, yet most current approaches rely on complex adapter and prompt mechanisms that increase tunable parameters. In this paper, we propose \textbf{\texttt{\textcolor{lora_blue}{Point}\textcolor{lora_orange}{LoRA}}}, a simple yet effective method that combines low-rank adaptation (LoRA) with multi-scale token selection to efficiently fine-tune point cloud models. Our approach embeds LoRA layers within the most parameter-intensive components of point cloud transformers, reducing the need for tunable parameters while enhancing global feature capture. Additionally, multi-scale token selection extracts critical local information to serve as prompts for downstream fine-tuning, effectively complementing the global context captured by LoRA. The experimental results across various pre-trained models and three challenging public datasets demonstrate that our approach achieves competitive performance with only 3.43\% of the trainable parameters, making it highly effective for resource-constrained applications. 
Source code is available at: \href{https://github.com/songw-zju/PointLoRA}{https://github.com/songw-zju/PointLoRA}.
\end{abstract}

%% file: sections/1_intro.tex
\section{Introduction}
\label{sec:intro}
3D point cloud learning plays a vital role in computer vision, advancing the understanding and reconstruction of complex 3D scenes~\cite{guo2020deep, bello2020deep, xiao2024survey,li2024is,kong2023rethinking}. Training deep neural networks on point clouds presents unique challenges due to their unordered structure, sparsity, and irregularity. In recent years, point-based methods~\cite{qi2017pointnet, qi2017pointnet++, li2018pointcnn, qian2022pointnext, liang2024pointmamba, xie2020pointcontrast} have made significant progress in addressing these challenges. 

As the volume of available point cloud data grows, there has been a surge in interest toward pre-training on unlabeled point clouds to learn generalizable representations~\cite{yu2022point, xie2020pointcontrast, qi2023recon, zheng2024point,sautier2022slidr,liu2023seal}. Robust pre-trained models enable efficient fine-tuning for downstream tasks, reducing dependency on labeled data, accelerating convergence, and improving accuracy \cite{kong2023robo3d,xu2024superflow}. However, conventional full fine-tuning can disrupt pre-trained knowledge, potentially diminishing the model’s generalization capability. Moreover, full fine-tuning requires storing multiple versions of model weights, leading to substantial storage costs as datasets and tasks expand, along with increased computational demands for larger pre-trained models.

Recent studies~\cite{zha2023instance, zhou2024dynamic, zhang2024positional, tang2024point} have made strides in introducing parameter-efficient fine-tuning (PEFT) methods to pre-trained point cloud models, building on approaches widely used in natural language processing (NLP) and 2D vision~\cite{ding2023parameter, xin2024parameter}. As an illustration, IDPT~\cite{zha2023instance} utilizes instance-aware dynamic prompt tuning to enhance model robustness in downstream transfer learning tasks. The subsequent methods~\cite{zhou2024dynamic, tang2024point} combine prompt tuning with adapter tuning, while relying on carefully crafted adapters and specialized prompt design. 
PPT~\cite{zhang2024positional} presents positional prompt tuning, which doubles the sequence length and significantly increases computational requirements.

\begin{figure*}
    \centering
    \includegraphics[width=0.92\linewidth]{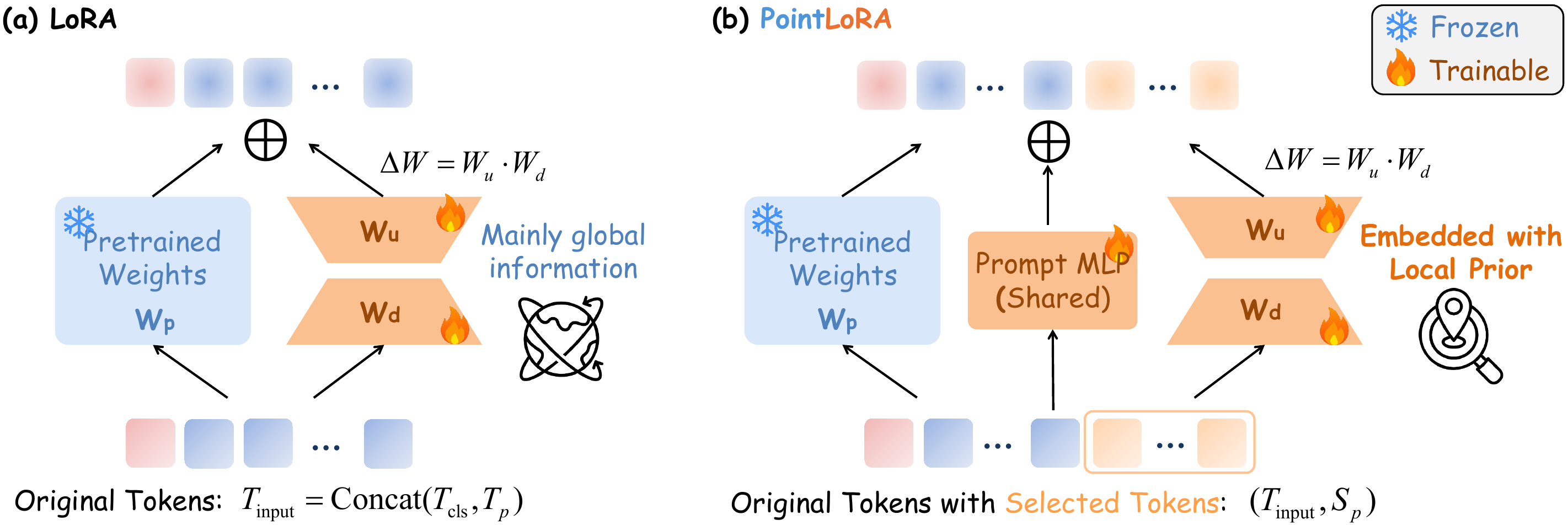}
    \vspace{-0.3cm}
    \caption{Comparing our proposed \textbf{\texttt{\textcolor{lora_blue}{Point}\textcolor{lora_orange}{LoRA}}} approach against vanilla LoRA methods. Both LoRA and our approach incorporate low-rank adaptation matrices into the pre-trained weights to extract global information from the point cloud sequence. Furthermore, our approach integrates tokens selected at various scales to capture local information, which is refined using a shared Prompt MLP and then output alongside the results derived from the original low-rank matrices.} 
    \label{fig:intro}
    \vspace{-5mm}
\end{figure*}

To fine-tune in a more efficient manner and further reduce the parameter count, we propose \textbf{\texttt{\textcolor{lora_blue}{Point}\textcolor{lora_orange}{LoRA}}}, which leverages low-rank adaptation (LoRA)~\cite{hu2021lora}, widely studied in large language models, to enable better fine-tuning for point cloud models. 
As shown in \cref{fig:intro}~(a), during LoRA fine-tuning, instead of directly updating the pre-trained weights, two auxiliary low-rank matrices, $W_u$ and $W_d$, are injected to update the network with fewer tunable parameters. From a 3D learning perspective, these matrices function as fully connected layers that effectively capture global point cloud features but lack local information extraction, akin to the architecture of PointNet \cite{qi2017pointnet}.

Unlike 2D visual or language data, 3D point clouds contain distinctive local geometric features critical for downstream tasks. 
Additionally, not all local positions contribute equally to the model, as their importance varies depending on the specific task and data characteristics. 
To address this, we further design a \textbf{Multi-Scale Token Selection} module, providing local prior information as a prompt to complement LoRA’s global feature aggregation capabilities. 
Using multi-scale sampling and $k$-nearest neighbor aggregation, we derive tokens that represent features across different scales. A shared Mask Predictor then selects a subset of these tokens to act as prompts. 
During fine-tuning, tokens from various positions are dynamically selected, processed through a \textbf{Prompt MLP}, and incorporated into the LoRA layer, as illustrated in \cref{fig:intro}~(b). This design enables the integration of essential local geometric features with global context, allowing the pre-trained model to adapt effectively to diverse downstream tasks.

Our main contributions are summarized as follows.
\begin{itemize}
    \item We propose \textbf{\texttt{\textcolor{lora_blue}{Point}\textcolor{lora_orange}{LoRA}}}, a simple yet effective scheme for parameter-efficient fine-tuning in point cloud learning.
    \item Low-rank adaptation is injected into the most parameter-intensive components of the point cloud transformer to capture global information, with a corresponding analysis in the context of PointNet provided.
    \item We further design a multi-scale token selection module that dynamically incorporates local geometric information as prompt, enriching global low-rank adaptation with the crucial local context.
    \item Extensive experiments across three challenging datasets and two widely used pre-trained models validate the effectiveness of our method, achieving state-of-the-art performance with only $3.43\%$ of the trainable parameters.
\end{itemize}

%% file: sections/2_related_work.tex
\section{Related Work}
\label{sec:related_work}

\noindent \textbf{Model Pre-Training on Point Clouds.} 
Self-supervised pre-training for 3D point clouds has been widely explored \cite{yu2022point, xie2020pointcontrast, qi2023recon, zheng2024point,sautier2022slidr,liu2023seal, wang2022meta, wang2023lidar2map}, generally categorized into contrastive-based approaches \cite{chen2023clip2scene, afham2022crosspoint, xie2020pointcontrast, nunes2022segcontrast,xu2024superflow} and masked signal reconstruction methods \cite{pang2022masked, zhang2022point, chen2024pointgpt, yu2022point, chen2023masked}. In contrastive learning, PointContrast~\cite{xie2020pointcontrast} and CrossPoint~\cite{afham2022crosspoint} leverage different views or instances to extract latent features through contrasting representations. SegContrast~\cite{nunes2022segcontrast} extends this approach to outdoor LiDAR data, extracting class-agnostic segments from augmented views \cite{kong2023lasermix}. Reconstruction-based methods, such as Point-BERT~\cite{yu2022point} and Point-MAE~\cite{pang2022masked}, employ masked point cloud reconstruction for pre-training. In ACT~\cite{dong2022autoencoders}, a pre-trained autoencoder serves as a cross-modality teacher, guiding 3D point cloud learning through knowledge distillation~\cite{hinton2015distilling, wang2024not, wang2024label}. PointGPT~\cite{chen2024pointgpt} adapts the GPT framework~\cite{mann2020language} for auto-regressive generation on point clouds. Additionally, Qi~\etal~\cite{qi2023recon, qi2024shapellm} propose a combination of reconstruction and cross-modal contrastive learning within generative models. While full fine-tuning is commonly applied to these pre-trained models for downstream tasks, the high computational costs and potential dilution of pre-trained knowledge motivate our exploration of efficient fine-tuning strategies.

\noindent \textbf{Parameter-Efficient Fine-Tuning (PEFT).} 
PEFT methods have gained traction in NLP and 2D vision tasks~\cite{ding2023parameter, shi2024dept, zhang2023adaptive, chen2022adaptformer, jia2022visual, houlsby2019parameter, li2021prefix, xiao2024event} to enhance downstream performance with minimal tunable parameters. Mainstream PEFT approaches include prompt and prefix tuning, adapter-based methods, and low-rank adaptation. Prompt and prefix tuning~\cite{lester2021power, li2021prefix, jia2022visual, shi2024dept} introduce additional tokens as tunable components, with VPT \cite{jia2022visual} pioneering the use of prompt tokens for pre-trained vision transformers (ViTs) \cite{dosovitskiy2020image,yang2025kolmogorovarnold, song2022transformer, li2023dropkey}. Adapter-based tuning inserts trainable modules between frozen layers \cite{houlsby2019parameter, chen2022adaptformer, sung2022lst, li2024adapter}, while LoRA~\cite{hu2021lora} applies a low-rank approximation to update linear layers within attention blocks. Building on these techniques, recent innovations explore alternative adapter placements~\cite{lian2022scaling, he2021towards}, tuning bias terms~\cite{zaken2022bitfit}, and combining with Mixture-of-Experts (MoE) techniques \cite{zadouri2023pushing, liu2023moelora}.
Recent studies have also adapted PEFT to point cloud analysis~\cite{zha2023instance, zhou2024dynamic, zhang2024positional, tang2024point, liang2024parameter}. IDPT \cite{zha2023instance} extends prompt tuning to DGCNN \cite{wang2019dynamic} for instance-aware prompt extraction. Point-PEFT~\cite{tang2024point} and DAPT~\cite{zhou2024dynamic} combine prompt tuning with adapters to further improve fine-tuning performance, while PPT~\cite{zhang2024positional} introduces positional prompt tuning for efficient 3D representation learning. In contrast to these approaches, our work uniquely integrates low-rank adaptation with multi-scale token selection, achieving competitive performance with a significantly reduced parameter footprint.

%% file: sections/3_prelim.tex
\section{Preliminary}
\label{sec:preli}
In this section, we revisit the architecture of point cloud transformer and the corresponding fine-tuning prototypes.
\subsection{Point Cloud Transformer}
\noindent \textbf{Point Tokenizer.} 
\label{sec:pt}
Due to the sparsity and irregularity of point clouds, it is necessary to transform them into a token sequence suitable for processing by the transformer.
We follow standard settings~\cite{yu2022point, pang2022masked} by segmenting the point cloud into irregular patches using the farthest point sampling ($\operatorname{FPS}$) and the $k$-nearest neighbors ($k$-NN) algorithm. 
Formally, given the point cloud $P = \{p_1, p_2, \ldots, p_N\} \in \mathbb{R}^{N\times3}$ with $N$ points, the $g$ group centers are obtained with $\operatorname{FPS}$:
\begin{equation}
C_g=\operatorname{FPS}\left(P\right), \quad C_g \in \mathbb{R}^{g \times 3},
\end{equation}
where $C_g$ is the selected center points. Then, we use $k$-NN to select $k$ nearest neighbor points for each center in $C_g$:
\begin{equation}
N_p=k\text{-NN}\left(P, C_g\right), \quad N_p \in \mathbb{R}^{g \times k \times 3},
\end{equation}
where $N_p$ is the corresponding neighbor point patch. 
Notably, these local point patches are made unbiased by subtracting their center points, which promotes better convergence.
Then a $\operatorname{mini-PointNet}$~\cite{qi2017pointnet} is utilized to embed the point patches into discrete tokens $T_p$:
\begin{equation}
T_p=\operatorname{mini-PointNet}\left(N_p\right), \quad T_p \in \mathbb{R}^{g \times d},
\end{equation}
where $d$ is the embedding dimension.
Following the tokenization process, the 3D point cloud is converted into a feature vector, enabling subsequent processing with Transformer architectures similar to those used in natural language processing and 2D vision.

\noindent \textbf{Transformer Block.} 
With a classification token $T_{\text{cls}}$ and the obtained $T_p$, the input $T_{\text{input}}=\operatorname{Concat}(T_{\text{cls}}, T_p)$ is further processed by $L$-layer transformer blocks. In each block, we first project $T_{input}$ into query, key, and value spaces:
\begin{equation}
\mathbf{Q} = T_{\text{input}} \mathbf{W}^Q, \quad \mathbf{K} = T_{\text{input}} \mathbf{W}^K, \quad \mathbf{V} = T_{\text{input}} \mathbf{W}^V,
\end{equation}
where $\mathbf{W}^Q, \mathbf{W}^K, \mathbf{W}^V \in \mathbb{R}^{d \times d}$ are trainable matrices in \texttt{qkv projection} layer. 
The attention layer then computes the self-attention with a skip connection as follows:
\begin{equation}
T^{\prime}_{\text{output}}=\operatorname{Attention}\left(\mathbf{Q}, \mathbf{K}, \mathbf{V}\right)+ T_{\text{input}}.
\end{equation}
Then, the final output $T_{\text{output}}$ is computed as:
\begin{equation}
T_{\text{output}}=\texttt{FFN}\left(\operatorname{LN}\left(T^{\prime}_{\text{output}}\right)\right)+T^{\prime}_{\text{output}},
\end{equation}
where $\operatorname{LN}$ is the layer norm function~\cite{ba2016layer} and $\operatorname{FFN}$ is the feed-forward network layer~\cite{bebis1994feed}. 
Through $L$ Transformer blocks, the network extracts deep and essential features from the discrete tokens.
Notably, the \texttt{qkv projection} parameters within the attention layer, along with those in the $\texttt{FFN}$ layer constitute a significant portion of the model’s parameters and are crucial for performance.

\subsection{Fine-Tuning Prototype} 
Leveraging well-pretrained models, fine-tuning prototypes to adapt these models to diverse downstream tasks can be categorized as follows.

\noindent \textbf{Full Fine-Tuning.} 
Full fine-tuning is the predominant approach in current point cloud pre-training research for downstream tasks, involving the direct loading of pre-trained model parameters and joint training of the encoder alongside the task-specific head. 
Although this method offers a theoretically higher accuracy potential, it also entails substantial computational and storage overhead.

\noindent \textbf{Parameter-Efficient Fine-Tuning.}
Parameter-efficient fine-tuning adapts pre-trained models to downstream tasks by freezing most of the model’s parameters, introducing lightweight modules, and selectively fine-tuning a small subset of parameters. 
This approach provides a more practical solution, particularly as model sizes continue to increase.
Existing methods typically combine adapter tuning, which inserts bottleneck-like layers, and prompt tuning, which introduces prior knowledge, to fine-tune point cloud pre-trained models. 

%% file: sections/4_method.tex
\begin{figure*}[t]
    \centering
    \includegraphics[width=1.0\linewidth]{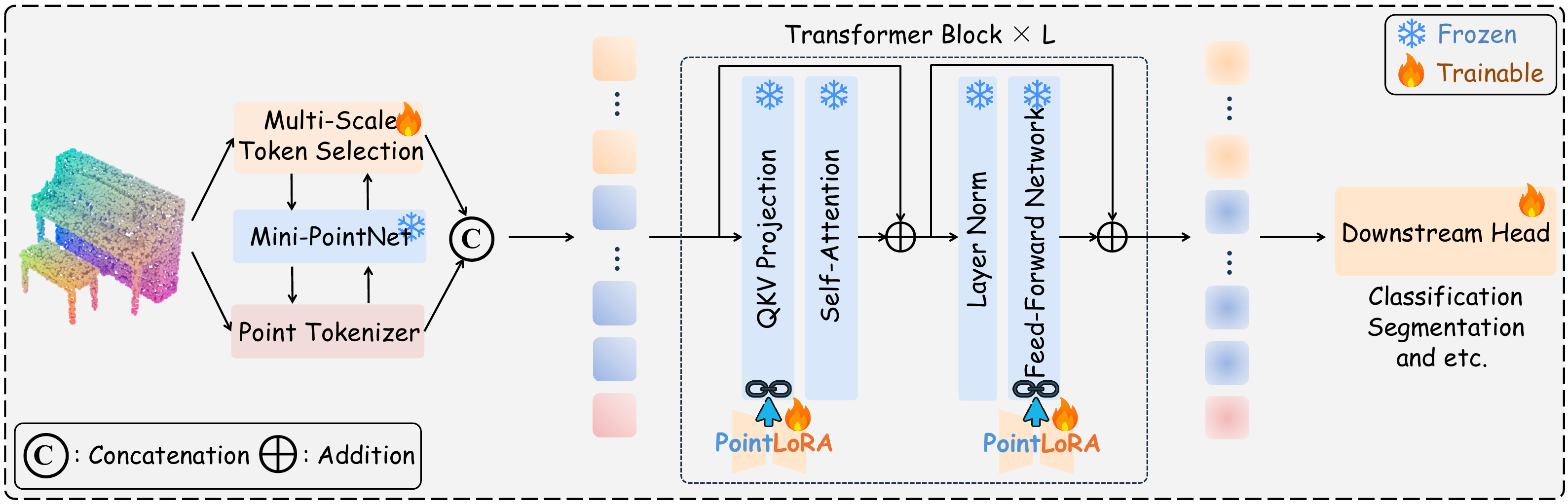}
    \vspace{-0.6cm}
    \caption{Overview of \textbf{\texttt{\textcolor{lora_blue}{Point}\textcolor{lora_orange}{LoRA}}} integrated into point cloud transformer pipeline. Given an input point cloud, we first tokenize it using the original {Point Tokenizer} and perform token selection across multiple scales ({Multi-Scale Token Selection}). The tokens from both components are then concatenated and fed into the {Transformer Block}. Our approach is injected into the \texttt{qkv projection} and \texttt{FFN} layers, utilizing a shared Prompt MLP within these layers to enhance parameter efficiency.}
    \label{fig:framework}
\end{figure*}

\section{PointLoRA}
\label{sec:method}
\subsection{Overview} 
In this work, we aim to provide a simple but effective parameter-efficient fine-tuning method for point cloud pre-trained models.
The complete framework of our proposed \textbf{\texttt{\textcolor{lora_blue}{Point}\textcolor{lora_orange}{LoRA}}}, integrated into the existing point cloud transformer, is illustrated in \cref{fig:framework}, addressing the parameter-intensive nature of the transformer-based network design.
Our approach can effectively enhance downstream task performance while keeping most of the pre-trained model parameters frozen.

In the following section, we first outline the establishment of the vanilla LoRA baseline and analyze the factors contributing to its effectiveness in point cloud learning. Then we present how our method extends LoRA’s feature extraction capabilities to achieve competitive performance in 3D scenarios.

\subsection{Vanilla LoRA Baseline}
As mentioned in \cref{sec:pt}, the \texttt{qkv projection} and feed-forward network (\texttt{FFN}) layers in point cloud transformer significantly increase the number of parameters, limiting the fine-tuning efficiency and hindering deployment in resource-limited environments. 
To address this issue, we incorporate Low-Rank Adaptation (LoRA) into these layers by introducing low-rank matrices to adapt pre-trained weights, thereby reducing the number of trainable parameters while preserving model performance.
Given a pre-trained weight matrix $W_p$, LoRA modifies it as follows:
\begin{equation}
W_{\text{update}} = W_p + \Delta W =  W_p + W_u \cdot W_d,
\end{equation}
where $\Delta W$ and $W_{\text{update}}$ denotes the updated weight and new weight matrix, respectively. 
$W_u \in \mathbb{R}^{d \times r}$ and $W_d \in \mathbb{R}^{r \times d}$ are low-rank matrices, where $r \ll d$ indicates the rank, controlling the complexity of the adaptation.

During training, only $W_u$ and $W_d$ are updated, while the pre-trained weight $W_p$ remains frozen. 
For inference, the adaptation $\Delta W$ can be merged with the original weights, yielding in a single consolidated weight matrix:
$W_{\text{infer}} = W_p + \Delta W$.
This consolidation enables the model to retain the benefits of low-rank adaptation while avoiding additional computational overhead in the inference phase.

\noindent \textbf{Analysis.} 
LoRA’s architecture is particularly well-suited for point cloud data, as it can capture the complex relationships between points essential to understanding 3D shapes and spatial arrangements. 
Its MLP-like structure (\textit{i.e.}, two low-rank matrices) effectively learns global features while aligning with PointNet’s principles of handling unordered point sets and extracting permutation-invariant features. 
The synergy between LoRA and PointNet’s feature extraction capabilities establishes a robust framework for harnessing the rich information in point cloud data, enhancing both generalization and adaptability in complex environments.

\subsection{PointLoRA with Token Selection}
While global features provide a comprehensive understanding of the overall structure of the point cloud, local features capture intricate details vital for accurate downstream task performance.
To fully investigate LoRA’s capabilities, it is crucial to complement the extracted global features with robust local features.
The tokenization process produces a sequence of tokens that capture various aspects of the point cloud data. However, not all tokens are valuable for downstream tasks; some may encapsulate redundant or non-informative content, offering limited contribution to overall model performance.
This necessitates a systematic approach for filtering and selecting informative tokens to enhance the model's representation of local features.

\begin{figure}
    \centering
    \includegraphics[width=0.95\linewidth]{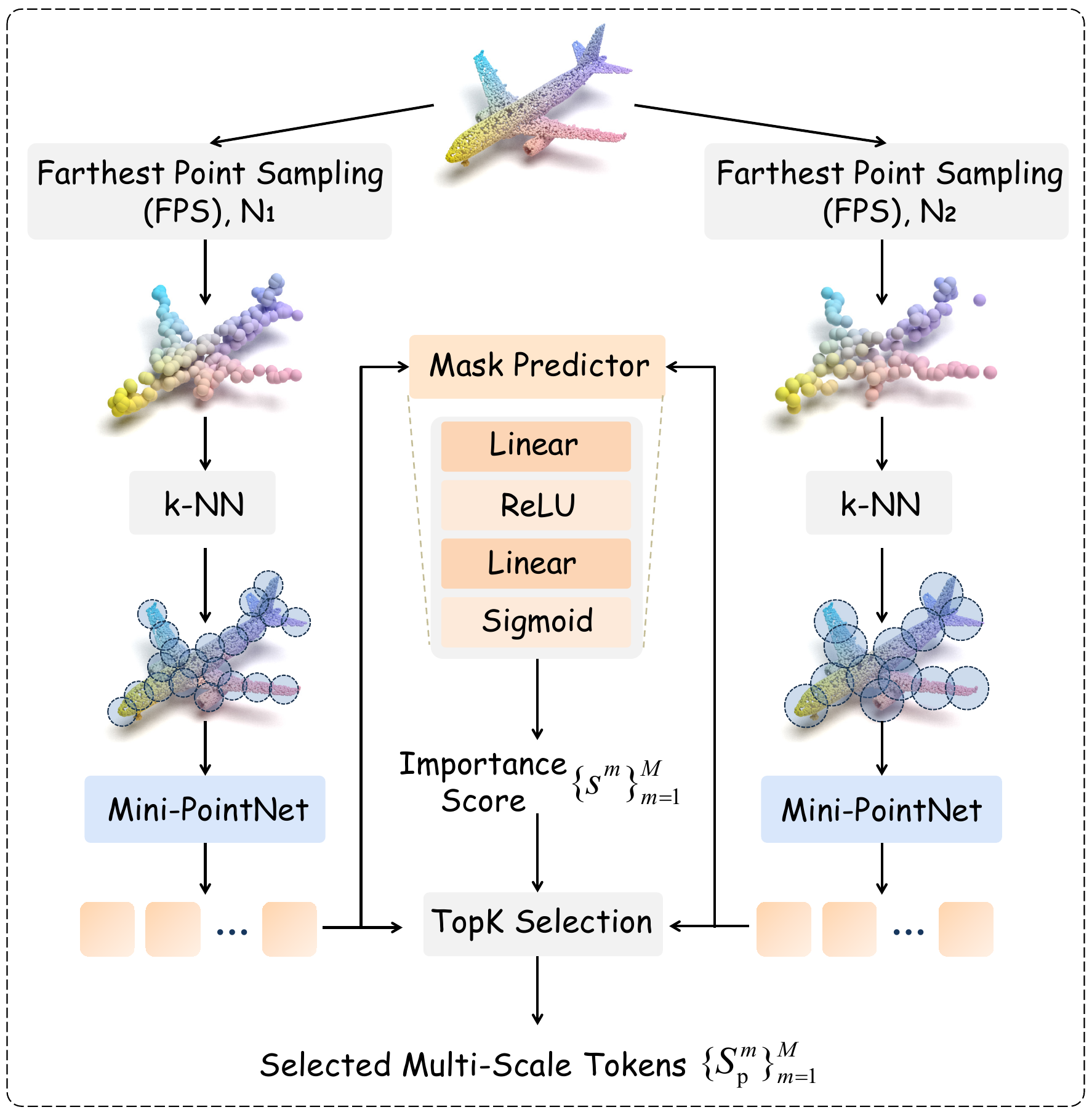}     
    \vspace{-0.2cm}
    \caption{Illustration of \textbf{Multi-Scale Token Selection}. In a two-scale setup, we first sample different numbers of center points, then cluster around each center point and apply $\operatorname{Mini-PointNet}$ encoding to generate the corresponding tokens. These tokens are also fed into a Mask Predictor to estimate importance scores, allowing us to select the Top-K tokens at each scale.}
    \label{fig:msts}
\end{figure}

To tackle the above concerns, we introduce a \textbf{Multi-Scale Token Selection} module integrated with vanilla LoRA, which is presented in \cref{fig:msts}. This module selects tokens from the raw point cloud at multiple scales, allowing the model to capture local features at different levels of detail. 
Additionally, \textbf{\texttt{\textcolor{lora_blue}{Point}\textcolor{lora_orange}{LoRA}}} incorporates a shared Prompt MLP to further embed local information from these selected tokens, enhancing the vanilla LoRA’s performance.

\noindent \textbf{Multi-Scale Token Generation.} Specifically, we apply farthest point sampling ($\operatorname{FPS}$) on the input point cloud $P$ with varying numbers of centroids (\textit{i.e.}, $N_1, N_2, ..., N_M$) to generate tokens at $M$ different scales.
As described in \cref{sec:preli}, we can obtain $M$ sets of center points, $C_g^1, C_g^2, \dots, C_g^M$, each containing $g_1, g_2, \dots, g_M$ center points. 
Subsequently, $k$-NN is employed to retrieve the corresponding neighbor point patches, which are then embedded into discrete tokens using a shared \( \operatorname{Mini-PointNet} \). 

\noindent \textbf{Mask Predictor for Selection.} 
The above process yields $M$ token sets, $T_p^1, T_p^2, \dots, T_p^M$, encoding local information at multiple scales.
Our goal is to select tokens from these sets that are most beneficial for fine-tuning on downstream tasks.
Therefore, a simple but effective Mask Predictor is designed to estimate the importance of each token, producing a score vector $s^m \in \mathbb{R}^{g_m\times1}$ for each token in $T_p^m \in \mathbb{R}^{g_m \times d}, m=1,...,M$.
In particular, the Mask Predictor comprises two multi-layer perceptron (MLP) layers followed by a Sigmoid activation function:
\begin{equation}
s^m = \operatorname{Sigmoid}(\operatorname{MLP}((T_p^m))).
\end{equation}
This design enables the model to assign scores between $0$ and $1$ to each token, indicating their relative importance based on structural cues.
We then select a predefined number $N_1^{\prime}, ..., N_M^{\prime}$ of tokens from each scale based on their importance scores using Top-K selection, resulting in a refined token set $\{S_{\text{p}}^{m}\}_{m=1}^{M}$:
\begin{equation}
S_{\text{p}}^{m} = \operatorname{TopK}(T_p^m, N_m^{\prime}), m=1,..., M.
\end{equation}

\noindent \textbf{Local Geometry Prompt.} 
With the selected $N_s = N_1^{\prime}+...+N_M^{\prime}$ tokens $S_p$,  we can capture crucial local information for downstream tasks.
These tokens are then concatenated with the original inputs to form a local geometry prompt.
In each LoRA layer of the point cloud transformer block, a Prompt MLP with a GELU activation function~\cite{hendrycks2016gaussian} encodes the new input features alongside the local prompt. 
This process integrates the encoded features of the selected tokens with the LoRA adaptation output. 
The resulting combined output can be formulated as follows:
\begin{equation}
O_{\text{update}} = \text{Prompt MLP}(T_{\text{input}}, S_p) + \Delta W \cdot (T_{\text{input}}, S_p),
\end{equation}
where $O_{\text{update}}$ is the updated amount of the \textbf{\texttt{\textcolor{lora_blue}{Point}\textcolor{lora_orange}{LoRA}}} layer.
Notably, to further optimize the efficiency of the parameters, the Prompt MLP is configured independently for both the \texttt{qkv projection} and \texttt{FFN} layers, while being shared across all blocks.

The multi-scale token selection process improves model performance by prioritizing compact yet informative local features during fine-tuning. 
By incorporating this approach, our scheme effectively refines both global and local feature representations, thereby enhancing its adaptability and accuracy across a range of 3D tasks.

\subsection{Training \& Inference}
\noindent \textbf{Overall Loss Function.} 
When fine-tuning on diverse downstream tasks across various datasets, we follow the common practice of maintaining consistency between the adopted loss and the loss function of the original task.
Additionally, we introduce a regularization term to supervise the Mask Predictor. In general, the total loss function ${\mathcal{L}_{\text{total}}}={\mathcal{L}}_{\text{task}}+\lambda \cdot {\mathcal{L}}_{\text{mask}},$
where ${\mathcal{L}}_{\text{task}}$ is the task-specific loss for classification or segmentation. ${\mathcal{L}}_{\text{mask}}$ is the regularization loss for mask predictor with the balanced weight $\lambda$:
\begin{equation}
\begin{aligned}
\mathcal{L}_{\text{mask}} = & -\frac{1}{N_\text{total}} \sum\nolimits_{i=1}^{N_\text{total}} ( s_i \log(s_i + \epsilon) + \\& (1 - s_i) \log(1 - s_i + \epsilon) ),
\end{aligned}   
\end{equation}
where $N_\text{total}=N_1+N_2...+N_M$, $s_i$ denotes the importance score for each token and $\epsilon$ is a small constant (\textit{e.g.}, $10^{-6}$) to prevent undefined values in the logarithm.

\noindent \textbf{Inference.} 
During inference, the additional parameters introduced by LoRA in the transformer block are directly merged into the original pre-trained weights, with the only newly added component being a small shared MLP dedicated to embed the selected tokens.

%% file: sections/5_exps.tex
\input{tables/joint_tab}

\section{Experiments}
\label{sec:exps}
\subsection{Setup}

\noindent \textbf{Datasets.}  
We validate the effectiveness of our approach through extensive experiments on three widely used 3D datasets: ScanObjectNN~\cite{uy2019revisiting}, ModelNet40~\cite{wu20153d}, and ShapeNetPart~\cite{yi2016scalable}.
ScanObjectNN~\cite{uy2019revisiting} is a challenging real-world 3D object classification dataset, which contains approximately $15,000$ indoor point cloud instances across $15$ categories.
The object classification task is performed on three variants of increasing complexity, OBJ-BG, OBJ-ONLY, and PB--T50--RS, each representing different levels of scene realism and occlusion.
ModelNet40~\cite{wu20153d}, a classical synthetic dataset for 3D object recognition, consists of $12,311$ meshed 3D CAD objects that span $40$ categories.
We perform synthetic object classification and few-shot learning experiments on ModelNet40 to evaluate the robustness of the model in both standard and low-resource settings.
Additionally, to evaluate performance on detailed structures and component segmentation, we provide experiments on ShapeNetPart~\cite{yi2016scalable}, a widely used benchmark for part segmentation, comprising $16,881$ point-level synthetic objects across $16$ object categories and $50$ part categories. 

\noindent \textbf{Evaluation Metrics.}
In the evaluation, overall accuracy (OA) is adopted to assess performance on the 3D object classification task, representing the ratio of correctly classified instances to the total number of instances, thus providing an aggregate score over all classes.
For the part segmentation task, we employ mean Intersection over Union (mIoU) to evaluate the overlap between prediction and ground-truth segments, averaged across all classes.

\begin{figure*}
    \centering
    \includegraphics[width=0.95\linewidth]{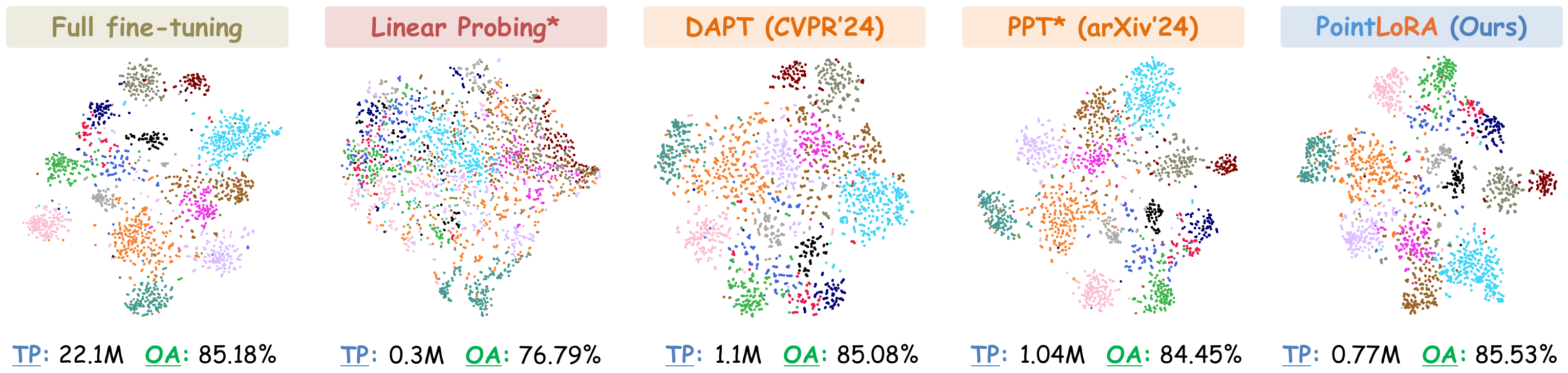}
    \vspace{-0.2cm}
    \caption{The t-SNE visualization results on the PB-T50-RS split of the ScanObjectNN dataset \cite{uy2019revisiting} with different fine-tuning schemes. We adopt Point-MAE~\cite{pang2022masked} as the baseline model for fair comparison. \textbf{TP:} Number of tunable parameters. \textbf{OA:} Overall accuracy. Symbol $*$ denotes re-produced with official implementation. Best viewed in colors and zoomed-in for additional details.} 
    \label{fig:tsne}
    \vspace{-3mm}
\end{figure*}

\noindent \textbf{Implementation Details.}
Our method can be directly integrated into existing point cloud pre-trained models. 
During fine-tuning, we follow a commonly used setup, freezing most parameters of the pre-trained model, and primarily updating the newly inserted parameters. 
The number of scales, $M$, is set to $2$. At these two scales, we apply farthest point sampling (\(\operatorname{FPS}\)) and $k$-NN clustering with ($128$, $32$) and ($64$, $64$) center and neighboring points, respectively, and then select tokens with \(N_1^{\prime}=32\) and \(N_2^{\prime}=8\). 
The balanced weight $\lambda$ for Mask Predictor learning is set to $0.004$
For real-world and synthetic object classification, we fine-tune using Point-MAE~\cite{pang2022masked}, training for $300$ epochs with a learning rate of $5e^{-4}$ and a weight decay of $0.05$. For few-shot learning and part segmentation, we utilize the Recon~\cite{qi2023recon} model for fine-tuning to further validate the generalizability of our approach, with a learning rate of $2e^{-4}$ for part segmentation. 
All experiments are conducted on a single GPU.
More implementation details with different baselines are provided in the Supplementary Material.

\input{tables/few_shot}
\subsection{Comparative Study}
\noindent \textbf{Real-World \& Synthetic Object Classification.} 
We conduct real-world and synthetic object classification on three variants of ScanObjectNN (OBJ-BG, OBJ-ONLY, PB-T50-RS)~\cite{uy2019revisiting} and ModelNet40~\cite{wu20153d}, using default data augmentation and without voting. 
As illustrated in \cref{tab:joint_tab}, our proposed approach demonstrates consistent performance improvements, especially on the \textit{most challenging} variant, \textbf{PB-T50-RS}, where it is the only parameter-efficient fine-tuning method that surpasses the full fine-tuned Point-MAE~\cite{pang2022masked} ($85.53\%$ \vs $85.18\%$) and significantly outperforms the state-of-the-art method, DAPT~\cite{zhou2024dynamic}. 
Notably, our approach requires only $0.77$M tunable parameters, the fewest among all methods.
Furthermore, the t-SNE feature visualization~\cite{van2008visualizing} of existing methods and \textbf{\texttt{\textcolor{lora_blue}{Point}\textcolor{lora_orange}{LoRA}}} is presented in \cref{fig:tsne}.
Our method enables the pre-trained model to fine-tune with a minimal number of parameters, producing distinctive feature representations.

\noindent \textbf{Few-shot Learning.} 
We also perform few-shot learning experiments on ModelNet40~\cite{wu20153d} to validate the transfer learning capability of our method with limited annotations. 
Following standard protocols~\cite{he2022masked, qi2023recon}, we evaluate the fine-tuning performance with ReCon~\cite{qi2023recon} in the $5$-way / $10$-way and $10$-shot / $20$-shot settings, respectively. 
As shown in \cref{tab:fewshot}, our method achieves the best or second-best accuracy in all configurations, further demonstrating the generalizability of the proposed scheme.

\input{tables/segmentation}
\noindent \textbf{Point Cloud Part Segmentation.} 
The part segmentation results on the ShapeNetPart dataset \cite{yi2016scalable} are provided in \cref{tab:segmentation}. 
In this fine-grained scene understanding task, our approach still achieves competitive performance with a constrained parameter budget. Unlike prior classification models, the increase in the parameter count is primarily attributed to the segmentation head.

\noindent \textbf{Comparison with Other PEFT Methods.}
We further present a comparison with existing PEFT methods designed for other tasks, including NLP and 2D vision. 
We select the \textit{most challenging} \textbf{PB-T50-RS} variant~\cite{uy2019revisiting} and use the pre-trained Point-MAE model~\cite{pang2022masked} as a baseline.
As shown in \cref{tab:peft_compare}, these methods designed for other tasks provide limited performance gains when fine-tuned in 3D scenes, despite their parameter efficiency compared to existing 3D fine-tuning approaches. 
In contrast, our method not only achieves the highest accuracy, but also uses significantly fewer parameters than the three current fine-tuning techniques specifically developed for point clouds, underscoring the strong potential of \textbf{\texttt{\textcolor{lora_blue}{Point}\textcolor{lora_orange}{LoRA}}}.

\input{tables/peft_comp}

\subsection{Ablation Study}
In this section, we perform exhaustive ablation experiments on the challenging PB-T50-RS variant to investigate the rationale and effectiveness of the design choices of our proposed approach. The pre-trained Point-MAE~\cite{pang2022masked} is adopted as a baseline for a fair comparison.

\input{tables/abl_framework}

\noindent \textbf{Ablation on \textbf{\texttt{\textcolor{lora_blue}{Point}\textcolor{lora_orange}{LoRA}}} Scheme.}
Firstly, we provide the ablations on each module of the proposed method.
As illustrated in \cref{tab:abl_framework}, directly applying low-rank adaptation (LoRA) produces a significant improvement over linear probing, demonstrating the effectiveness of incorporating global information during the fine-tuning process.
Injecting selected tokens as prompts into LoRA further improves accuracy with only a marginal increase in the number of parameters. 
Finally, the local information obtained from selected tokens across multiple scales complements the global features of LoRA, achieving the best performance.

\noindent \textbf{Ablation on Low-Rank Adaptation.}
Then we perform ablation experiments on rank ($r$) in our approach, as shown in \cref{tab:abl_rank}. Different ranks have a noticeable impact on fine-tuning performance. Taking into account both parameter count and accuracy, we set the rank to $8$ to maximize the extraction of global information for fine-tuning.

\noindent \textbf{Ablation on Multi-Scale Token Selection.}
Given the critical improvement from multi-scale token selection for \textbf{\texttt{\textcolor{lora_blue}{Point}\textcolor{lora_orange}{LoRA}}}, we first perform ablation experiments on the number of tokens selected at the same scale. As shown in the upper part of \cref{tab:abl_token}, selecting $16$ tokens at the original scale yields the best results, demonstrating the effectiveness of token selection. 
Furthermore, when incorporating tokens from multiple scales, optimal performance is achieved by selecting $32$ and $8$ tokens from two different scales, as shown in the lower part of \cref{tab:abl_token}.

\input{tables/abl_rank}
\input{tables/abl_token}
\input{tables/abl_dim}

\noindent \textbf{Ablation on Prompt MLP.}
We provide the ablation study on the dimension of the shared Prompt MLP, as this affects the embedding of local information in our method. 
As illustrated in \cref{tab:abl_dim}, setting this dimension to $32$ yields the best fine-tuning performance. 
Additionally, adjusting the dimension causes only minimal changes in tunable parameter counts, suggesting flexibility in selecting this hyperparameter for different datasets and tasks.

%% file: tables/joint_tab.tex
\begin{table*}[ht]
    \footnotesize
    \setlength{\tabcolsep}{2.4mm}
    \centering
    \caption{
        Performance comparison on three variants of the ScanObjectNN~\cite{uy2019revisiting} and the ModelNet40~\cite{wu20153d} datasets, respectively, for real-world and synthetic object classification. Both the number of tunable parameters and overall accuracy (OA) are reported. All methods only employ the default data argumentation without voting as the baseline. $*$ denotes results reproduced from the public source code. 
    }
    \vspace{-0.2cm}
    \begin{tabular}{lrcccccc}
    
    \toprule
    \multirow{2.3}{*}{\textbf{Methods}} &\multirow{2.3}{*}{\textbf{Publication}} &\multirow{2.3}{*}{\textbf{Tunable Params.}}  &\multicolumn{3}{c}{\textbf{ScanObjectNN}} &\multicolumn{2}{c}{\textbf{ModelNet40}}\\
		\cmidrule(r){4-6} \cmidrule{7-8}
	&  & &OBJ-BG & OBJ-ONLY &PB-T50-RS & Points Num. & OA (\%)      \\
    \midrule\midrule
    \rowcolor{mygray1} \multicolumn{8}{c}{{\em Traditional Supervised Learning Only}} \\
    \midrule
    PointNet~\cite{qi2017pointnet} & CVPR'17 & $3.5$ M   & $73.3$  & $79.2$  & $68.0$ & 1k & $89.2$ \\
    PointNet++~\cite{qi2017pointnet++}   & NeurIPS'17 & $1.5$ M  & $82.3$  & $84.3$  & $77.9$ & 1k & $90.7$\\
    DGCNN~\cite{wang2019dynamic}  & TOG'19 & $1.8$ M  & $82.8$  & $86.2$  & $78.1$ & 1k & $92.9$ \\
    MVTN~\cite{hamdi2021mvtn}  & ICCV'21 & $11.2$ M  & -     & -     & $82.8$ & 1k &  $93.8$\\
    PointNeXt~\cite{qian2022pointnext}  & NeurIPS'22  & $1.4$ M  & -     & -  & $87.7$ & 1k & $94.0$\\
    PointMLP~\cite{ma2022rethinking}  & ICLR'22 &  $13.2$ M  & -    & -     & $85.4$  & 1k & $94.5$\\
    RepSurf-U~\cite{ran2022surface} & CVPR'22 & $1.5$ M &  -  & -    & $84.3$  & 1k  & $94.4$ \\
    ADS~\cite{hong2023attention} & ICCV'23   & -  &  - & -   & $87.5$ & 1k  & $95.1$ \\
    \midrule
    \rowcolor{mygray1}\multicolumn{8}{c}{\em Self-Supervised Representation Learning (Full Fine-Tuning)} \\
    \midrule
    OcCo~\cite{wang2021unsupervised} & ICCV'21 & $22.1$ M & $84.85$ & $85.54$ & $78.79$ & 1k & $92.1$ \\
    Point-BERT~\cite{yu2022point}  & CVPR'22 & $22.1$ M  & $87.43$ & $88.12$ &  $83.07$ & 1k & $93.2$ \\
    MaskPoint~\cite{liu2022masked} & ECCV'22 & $22.1$ M & $89.70$ & $89.30$ &  $84.60$ & 1k &  $93.8$ \\
    Point-MAE~\cite{pang2022masked}  & ECCV'22 & $22.1$ M & $90.02$ & $88.29$ & $85.18$ & 1k & $93.8$ \\
    Point-M2AE~\cite{zhang2022point}  & NeurIPS 22 & $15.3$ M  & $91.22$ & $88.81$ & $86.43$ & 1k & $94.0$ 
    \\
    ACT~\cite{dong2022autoencoders}   & ICLR'23 & $22.1$ M  & $93.29$ & $91.91$  & $88.21$ & 1k & $93.7$\\
    \recon~\cite{qi2023recon} & ICML'23 & $43.6$ M & $94.15$ & $93.12$  & $89.73$ & 1k & $93.9$  \\
    
    \midrule
    \rowcolor{mygray1} \multicolumn{8}{c}{\textit{Self-Supervised Representation Learning (Parameter-Efficient Fine-Tuning)}} \\
    
    \midrule
    Point-MAE~\cite{pang2022masked} (Full-FT)& ECCV'22 & $22.1$ M ($100\%$) & $90.02$ & $88.29$ & {$85.18$} & 1k & $93.2$ \\
    Point-MAE + IDPT~\cite{zha2023instance}& ICCV'23 & $1.7$ M ($7.69\%$) & {$91.22$}\dplus{$+1.20$} & {$90.02$}\dplus{$+1.73$}& $84.94$\dtplus{$-0.24$} & 1k & {$93.3$}{\dplus{$+0.1$}}  \\
    Point-MAE + DAPT~\cite{zhou2024dynamic} & CVPR'24 & {$1.1$} M (${4.97}\%$)  & {$90.88$}\dplus{$+0.86$} & {$90.19$}\dplus{$+1.90$} & {$85.08$}\dtplus{$-0.10$} & 1k & {$93.5$}{\dplus{$+0.3$}} \\
    Point-MAE + PPT$^*$~\cite{zhang2024positional} & arXiv'24 & {$1.04$} M ({$4.57$}\%)  & {$89.84$}\dtplus{$-0.18$} & {$88.98$}\dplus{$+0.69$} & {$84.45$}\dtplus{$-0.73$}  & 1k & {$93.2$}{\dplus{$+0.0$}} \\
    \rowcolor{lora_orange!10}Point-MAE + \textbf{\texttt{\textcolor{lora_blue}{Point}\textcolor{lora_orange}{LoRA}}} & \textbf{Ours} & $\mathbf{0.77}$ M ($\mathbf{3.43}\%$) & {$90.71$}\dplus{$+0.69$} & {$89.33$}\dplus{$+1.04$} & {$85.53$}\dplus{$+0.35$} &1k & {$93.3$}{\dplus{$+0.1$}}\\
    \bottomrule
    \end{tabular}%
      \label{tab:joint_tab}
\vspace{-3mm}
\end{table*}

%% file: tables/few_shot.tex
\begin{table}[!t]
  \centering
  \caption{
    Performance comparison on ModelNet40~\cite{wu20153d} for few-shot learning. We report the scores of the overall accuracy (\%) $\pm$ the standard deviation (\%) without voting. The top two highest accuracies are highlighted in \textbf{bold} and \underline{underline}, respectively.
    }
    \vspace{-0.2cm}
    \scriptsize
    \setlength{\tabcolsep}{0.1mm}
    \renewcommand\arraystretch{1.1}
    \begin{tabular}{lccccc}
    \toprule
   \multirow{2.3}{*}{\textbf{Methods}}&\multirow{2.3}{*}{\textbf{Publication}} & \multicolumn{2}{c}{\textbf{$\mathbf{5}$-way}} & \multicolumn{2}{c}{\textbf{$\mathbf{10}$-way}} \\
\cmidrule{3-6}  &        & $10$-shot & $20$-shot & $10$-shot & $20$-shot \\
    \midrule\midrule
    \rowcolor{mygray1}\multicolumn{6}{c}{\textit{Self-Supervised Representation Learning (Full Fine-Tuning)}} \\
    \midrule
    OcCo~\cite{wang2021unsupervised}& ICCV'21      & $94.0$$\pm$$3.6$& $95.9$$\pm$$2.3$ & $89.4$$\pm$$5.1$ & $92.4$$\pm$$4.6$ \\
    Point-BERT~\cite{yu2022point}  &  CVPR'22    & $94.6$$\pm$$3.1$ & $96.3$$\pm$$2.7$ & $91.0$$\pm$$5.4$ & $92.7$$\pm$$5.1$ \\
    MaskPoint~\cite{liu2022masked}  &   ECCV'22   & $95.0$$\pm$$3.7$ & $97.2$$\pm$$1.7$ & $91.4$$\pm$$4.0$ & $93.4$$\pm$$3.5$ \\
    Point-MAE~\cite{pang2022masked} &   ECCV'22   & $96.3$$\pm$$2.5$ & $97.8$$\pm$$1.8$ & $92.6$$\pm$$4.1$ & $95.0$$\pm$$3.0$ \\
    Point-M2AE~\cite{zhang2022point} &  NeurIPS'22    & $96.8$$\pm$$1.8$ & $98.3$$\pm$$1.4$ & $92.3$$\pm$$4.5$ & $95.0$$\pm$$3.0$ \\
    ACT~\cite{dong2022autoencoders}  & ICLR'23      & $96.8$$\pm$$2.3$ & $98.0$$\pm$$1.4$ & $93.3$$\pm$$4.0$ & $95.6$$\pm$$2.8$ \\
    \midrule
    \rowcolor{mygray1}\multicolumn{6}{c}{\textit{Self-Supervised Representation Learning (Efficient Fine-Tuning)}} \\
       \midrule
    \recon~\cite{qi2023recon} (Full-FT) & ICML'23      & $97.3$$\pm$$1.9$ & $98.9$$\pm$$3.9$ & $93.3$$\pm$$3.9$ & $95.8$$\pm$$3.0$ \\
    \recon + IDPT~\cite{zha2023instance} & ICCV'23   & \underline{$96.9$}$\pm$$2.4$ & $98.3$$\pm$$0.7$ & $\mathbf{92.8}$$\pm$$4.0$ & $95.5$$\pm$$3.2$\\
   \recon + DAPT~\cite{zhou2024dynamic} & CVPR'24 & $95.6$$\pm${$2.8$}  &  {$97.7$}$\pm${$1.6$} & {$91.9$}$\pm${$4.1$} & {$94.6$}$\pm$$3.5$  \\

   \recon + PPT~\cite{zhang2024positional} & arXiv'24 & $\mathbf{97.0}$$\pm${$2.7$}  &  \underline{$98.7$}$\pm$$1.6$ & {$92.2$}$\pm${$5.0$} & \underline{$95.6$}$\pm$$2.9$  \\
   \rowcolor{lora_orange!10} \recon + \textbf{\texttt{\textcolor{lora_blue}{Point}\textcolor{lora_orange}{LoRA}}}  & \textbf{Ours} & \underline{$96.9$}$\pm${$2.7$}  & $\mathbf{98.8}$$\pm$${1.2}$  & \underline{$92.7$}$\pm$$4.4$  & $\mathbf{95.8}$$\pm$$2.9$  \\
    \bottomrule
    \end{tabular}%
  \label{tab:fewshot}%
  \vspace{-3mm}
\end{table}%

%% file: tables/segmentation.tex
\begin{table}[t]
  \centering
  \caption{
    Performance comparison on the ShapeNetPart~\cite{yi2016scalable} for part segmentation. Both the mIoU for all classes (Cls.) and instances (Inst.) are provided. TP indicates the number of tunable parameters. $\dag$ means the resuluts reported from DAPT~\cite{zhou2024dynamic}. $*$ denotes the results reproduced from the official implementation.
    }
    \vspace{-0.2cm}
    \scriptsize
    \setlength{\tabcolsep}{0.6mm}
    \renewcommand\arraystretch{1.1}
    \begin{tabular}{lcccc}
    \toprule
    \textbf{Methods} & \textbf{Publication} & \textbf{TP} & \textbf{Cls. mIoU (\%)}  & \textbf{Inst. mIoU (\%)}  \\
    \midrule\midrule
    \rowcolor{mygray1}\multicolumn{5}{c}{\textit{Traditional Supervised Learning Only}} \\
    \midrule
    PointNet \cite{qi2017pointnet} & CVPR'17  &- & $80.39$ & $83.7$ \\
    PointNet++  \cite{qi2017pointnet++}  & NeurIPS'17 &-  & $81.85$ & $85.1$ \\
    DGCNN \cite{wang2019dynamic} & TOG'19 & - & $82.33$ & $85.2$ \\
    APES~\cite{wu2023attention} & CVPR'23& - & $83.67$ & $85.8$\\
    \midrule
    \rowcolor{mygray1}\multicolumn{5}{c}{\textit{ Self-Supervised Representation Learning (Full Fine-Tuning)}} \\
    \midrule
    OcCo \cite{wang2021unsupervised} & ICCV'21 & $27.09$ M & $83.42$ & $85.1$ \\
    MaskPoint \cite{liu2022masked} & ECCV'22 & - & $84.60$ & $86.0$ \\
    Point-BERT \cite{yu2022point} & CVPR'22 & $27.09$ M & $84.11$ & $85.6$ \\
    Point-MAE \cite{pang2022masked} & ECCV'22 & $27.06$ M & $84.19$ & $86.1$ \\ 
    ACT \cite{dong2022autoencoders} & ICLR'23 &  $27.06$ M & $84.66$ & $86.1$ \\
    \midrule
    \rowcolor{mygray1}\multicolumn{5}{c}{\textit{ Self-Supervised Representation Learning (Efficient Fine-Tuning)}} \\
    \midrule
    \recon~\cite{qi2023recon}~(Full-FT) & ICML'23 & $27.06$ M & $84.52$ & $86.1$ \\
    \recon + IDPT$^\dag$~\cite{zha2023instance} &  ICCV'23 & $5.69$ M & $83.66$  & $\mathbf{85.7}$ \\
    \recon + DAPT~\cite{zhou2024dynamic} & CVPR'24 & {$5.65$} M & $83.87$ & $\mathbf{85.7}$ \\
    \recon + PPT$^*$~\cite{zhang2024positional} & arXiv'24 & {$5.62$} M & \underline{$83.88$} & \underline{$85.4$}\\
    \rowcolor{lora_orange!10}\recon + \textbf{\texttt{\textcolor{lora_blue}{Point}\textcolor{lora_orange}{LoRA}}} & \textbf{Ours} & $5.63$ M & $\mathbf{83.98}$ & \underline{$85.4$} \\
    \bottomrule
    \end{tabular}
  \label{tab:segmentation}
  \vspace{-6mm}
\end{table}

%% file: tables/peft_comp.tex
\begin{table}[t]
    \scriptsize
    \renewcommand\tabcolsep{8pt}
    \renewcommand\arraystretch{1.1}
    \caption{
        Performance comparison with other parameter-efficient methods designed for NLP and 2D Vision tasks on the hardest variant of ScanObjectNN~\cite{uy2019revisiting}.
    }
    \vspace{-0.2cm}
    \centering
    \begin{tabular}{lccc}
    \toprule
     \textbf{Methods} & \textbf{Publication} & \textbf{TP} & \textbf{PB-T50-RS} \\
    \midrule\midrule
     Point-MAE~\cite{liu2022masked}  &ECCV'22 & $22.1$ M & {$85.18$}  \\
     Linear probing &- & $0.3$ M & $75.99$\\
     \midrule
      + Adapter~\cite{houlsby2019parameter}&ICML'19 & $0.9$ M & $83.93$ \\
      + Perfix tuning~\cite{li2021prefix}& ACL'21 & $0.7$ M & $77.72$  \\
      + BitFit~\cite{zaken2022bitfit} & ACL'21 & $0.3$ M & $82.62$    \\
      + LoRA$^\dag$~\cite{hu2021lora} & ICLR'22 & $0.9$ M &  $81.74$   \\
      + VPT-Deep~\cite{jia2022visual}&ECCV'22 & $0.4$ M &  $81.09$ \\
      + AdaptFormer~\cite{chen2022adaptformer} &NeurIPS'22 & $0.9$ M & $83.45$ \\
      + SSF~\cite{lian2022scaling} & NeurIPS'22  & $0.4$ M & $82.58$\\
      \midrule
      + IDPT~\cite{zha2023instance} &ICCV'23 & $1.7$ M & $84.94$\\
      + DAPT~\cite{zhou2024dynamic} & CVPR'24 & $1.1$ M& {$85.08$} \\
      + PPT$^*$~\cite{zhang2024positional} & arXiv'24 & $1.04$ M & $84.45$ \\
      \rowcolor{lora_orange!10}+ \textbf{\texttt{\textcolor{lora_blue}{Point}\textcolor{lora_orange}{LoRA}}} & \textbf{Ours} & $0.77$ M & {$\mathbf{85.53}$} \\
    \bottomrule
    \end{tabular}
    \label{tab:peft_compare}
     \vspace{-3mm}
\end{table}

%% file: tables/abl_framework.tex
\begin{table}[t]
    \scriptsize
    \renewcommand\tabcolsep{6.4pt}
    \renewcommand\arraystretch{1.1}
    \centering
    \caption{The impact of each module of our scheme. We provide the overall accuracy (\%) on the hardest variant of ScanObjectNN~\cite{uy2019revisiting} and corresponding tunable parameters (TP). ``MS-FPS'' indicates multi-scale furthest point sampling.}
    \vspace{-0.2cm}
    \begin{tabular}{ccc|cc}
    \toprule
    \textbf{ LoRA}  & \textbf{Token Selection}  &  \textbf{MS-FPS} &  \textbf{TP}  & \textbf{PB-T50-RS} \\
    \midrule\midrule
    \multicolumn{3}{c|}{Full Fine-Tuning}  &$22.1$ M & $85.18$ \\
    \multicolumn{3}{c|}{Linear Probing}  &$0.27$ M& $75.99$ \\
    \midrule
    \ding {52} &  &   & $0.53$ M & $83.83$ \\
    \ding {52}& \ding {52} &   & $0.77$ M& $84.91$ \\
    \rowcolor{lora_orange!10}\ding {52}  & \ding {52} & \ding {52}  & $0.77$ M & {$\mathbf{85.53}$}  \\
    
    \bottomrule
    \end{tabular}
    \label{tab:abl_framework}
    \vspace{-5mm}
\end{table}

%% file: tables/abl_rank.tex
\begin{table}[t] 
    \centering
    \caption{Ablation study on rank ($r$) of the proposed method.}
    \vspace{-0.2cm}
    \scriptsize
     \renewcommand\tabcolsep{10.2pt} 
     \renewcommand\arraystretch{1.1}
    \begin{tabular}{cccc}
        \toprule
       \textbf{ Rank $r$}& \textbf{TP} & \textbf{Ratio} & \textbf{PB-T50-RS} \\
        \midrule\midrule
        $4$ & $0.66$ M & $2.96\%$ & $84.28$  \\
        \rowcolor{lora_orange!10} $8$ & $0.77$ M  & $3.43\%$  & {$\mathbf{85.53}$} \\
        $16$ & $0.99$ M & $4.37\%$ & $85.15$ \\
        $32$ & $1.44$ M & $6.32\%$ & $84.87$  \\
        \bottomrule
    \end{tabular}
    \label{tab:abl_rank}
    \vspace{-3mm}
\end{table}

%% file: tables/abl_token.tex
\begin{table}[t]
    \centering
    \caption{
        Ablation study on scale number $M$ and selected token number $N_s$ in Multi-Scale Token Selection.
    }
    \vspace{-0.2cm}
    \scriptsize
    \renewcommand\tabcolsep{12.5pt}
    \renewcommand\arraystretch{1.1}
    \begin{tabular}{ccc}
        \toprule
         \textbf{Scale $M$} & \textbf{Token Num. $N_s$} & \textbf{PB-T50-RS} \\
        \midrule\midrule
          $1$ & $64$ &  $84.80$ \\
          $1$ & $32$ & $84.91$  \\
          $1$ & $16$ &  $85.22$ \\
          \midrule
           $2$ & $64$ \& $16$ &  $84.39$ \\
         \rowcolor{lora_orange!10} $2$ & $32$ \& $8$ & $\mathbf{85.53}$  \\
          $2$ & $16$ \& $4$ &  $84.49$ \\
          \midrule
           $3$ & $32$ \& $16$ \& $8$ & $85.05$  \\  
        \bottomrule
    \end{tabular}
    \label{tab:abl_token}
    \vspace{-3mm}
\end{table}

%% file: tables/abl_dim.tex
\begin{table}[t]
    \renewcommand\tabcolsep{9pt}
    \renewcommand\arraystretch{1.1}
    \centering
    \caption{Ablation study on the dimension of Prompt MLP.}
    \vspace{-0.2cm}
    \scriptsize
    \begin{tabular}{cccc}
        \toprule
        \textbf{Dimension} &  \textbf{TP}  & \textbf{Ratio} &\textbf{PB-T50-RS} \\
        \midrule\midrule
        $8$  & $0.68$ M & $3.03\%$ & $83.55$ \\
        $16$ & $0.71$ M & $3.17\%$ & $84.77$ \\
        \rowcolor{lora_orange!10} $32$  & $0.77$ M & $3.43\%$ & {$\mathbf{85.53}$}\\
        $64$  & $0.90$ M & $3.96\%$ & $85.25$ \\
        \bottomrule 
    \end{tabular}
    \label{tab:abl_dim}
    \vspace{-2mm}
\end{table}

%% file: sections/6_conclusion.tex
\section{Conclusion}
\label{sec:conclusion}
In this paper, a simple yet effective parameter-efficient fine-tuning approach named \textbf{\texttt{\textcolor{lora_blue}{Point}\textcolor{lora_orange}{LoRA}}}, is presented with low-rank adaptation and multi-scale token selection for point cloud models.
Low-rank adaptation efficiently reduces the parameter-heavy components of the point cloud transformer architecture while capturing global information. 
Multi-scale token selection effectively encodes essential local features as prompt, further enhancing the fine-tuning process. 
The integration of global and local information enables our approach to achieve state-of-the-art results on the most challenging dataset with a minimal number of tunable parameters while also delivering competitive performance across other datasets and models.

%% file: supp.tex
\appendix
\maketitlesupplementary

\setcounter{figure}{0}
\setcounter{table}{0}
\renewcommand{\thefigure}{A\arabic{figure}}
\renewcommand{\thetable}{A\arabic{table}}

In this supplementary material, we further present the following descriptions and experiments to elaborate the results and conclusions addressed in the main paper.

\begin{itemize}
    \item Section \ref{sec:supp_a}: Detailed implementation specifications;
    \item Section \ref{sec:supp_b}: Extended experimental results;
    \item Section \ref{sec:supp_c}: Additional discussions for limitations, future work and broader impacts;
    \item Section \ref{sec:supp_d}: License and consent for public resources.

\end{itemize}

\section{Detailed Implementation Specifications}
\label{sec:supp_a} 
\subsection{Point-MAE-based Fine-tuning}
We leverage the Point-MAE~\cite{pang2022masked} pretrained model to perform object classification experiments on real-world data (ScanObjectNN~\cite{uy2019revisiting}) and synthetic data (ModelNet40~\cite{wu20153d}). The training settings are described on the left of \cref{tab:training}, following the pioneering work~\cite{pang2022masked, zhou2024dynamic}. All experiments are conducted on a single GeForce RTX 3090 GPU.

\subsection{ReCon-based Fine-tuning}
Similarly, more recent Recon~\cite{qi2023recon} pre-trained model is used for few-shot learning experiments on ModelNet40~\cite{wu20153d} and part segmentation on ShapeNetPart~\cite{yi2016scalable}. The training settings are detailed in the right half of \cref{tab:training}, following the general configurations~\cite{pang2022masked,zhang2024positional}, with all training conducted on a single GPU.

\section{Extended Experimental Results}
\label{sec:supp_b}
\subsection{More Ablation Studies}
\label{sec:more_abl}
Here we provide additional ablation experiments, adhering to the same settings described in the main paper. Specifically, we utilize the Point-MAE~\cite{pang2022masked} pre-trained model and report fine-tuning results on the most challenging variant, PB-T50-RS, of ScanObjectNN~\cite{uy2019revisiting}.

\noindent \textbf{Ablation on multi-scale token selection.}
We first conduct ablation experiments on the number of center points and neighboring points in the multi-scale token selection process. 
As shown in \cref{tab:abl_msfps}, these parameters influence the amount of information encoded in the tokens, which subsequently affects token selection and fine-tuning performance. For the two scales, we set these values at $(128, 32)$ and $(64, 64)$, respectively.

\input{tables/supp_abl_msfps}

\input{tables/supp_abl_loss}

\input{tables/supp_abl_layer}

\noindent \textbf{Ablation on the loss weight for mask learning.}
We also perform an ablation study on the loss weight \( \lambda \) of \( {\mathcal{L}}_{\text{mask}} \), which controls the strength of regularization applied to the Mask Predictor. As illustrated in \cref{tab:abl_loss}, the incorporation of mask loss \( {\mathcal{L}}_{\text{mask}} \) improves the diversity and quality of the selected tokens, leading to improved classification accuracy. However, an excessively large $\lambda$ for \( {\mathcal{L}}_{\text{mask}} \) may overly constrain token selection, causing a slight performance drop. 
To achieve the best accuracy, we set \( \lambda \) to \( 0.004 \).

\noindent \textbf{Ablation on the injected blocks for PointLoRA.}
Following DAPT~\cite{zhou2024dynamic}, we also experimented with injecting the designed components into only a subset of point cloud transformer blocks (\(L=12\) in total) to further reduce the number of tunable parameters. 
As shown in \cref{tab:insert}, limiting injections to shallow or deeper blocks results in varying degrees of performance degradation. 
This could be attributed to the fact that different blocks in the pre-trained model capture critical information related to distinct aspects of the input point cloud. Consequently, we choose to integrate \textbf{\texttt{\textcolor{lora_blue}{Point}\textcolor{lora_orange}{LoRA}}} into the \texttt{qkv projection} and \texttt{FFN} layers of all blocks, leaving the investigation of block-specific configurations for future research.

\input{tables/supp_training}

\subsection{Part Segmentation Visualization}

\begin{figure*}[t]
    \centering
    \includegraphics[width=\linewidth]{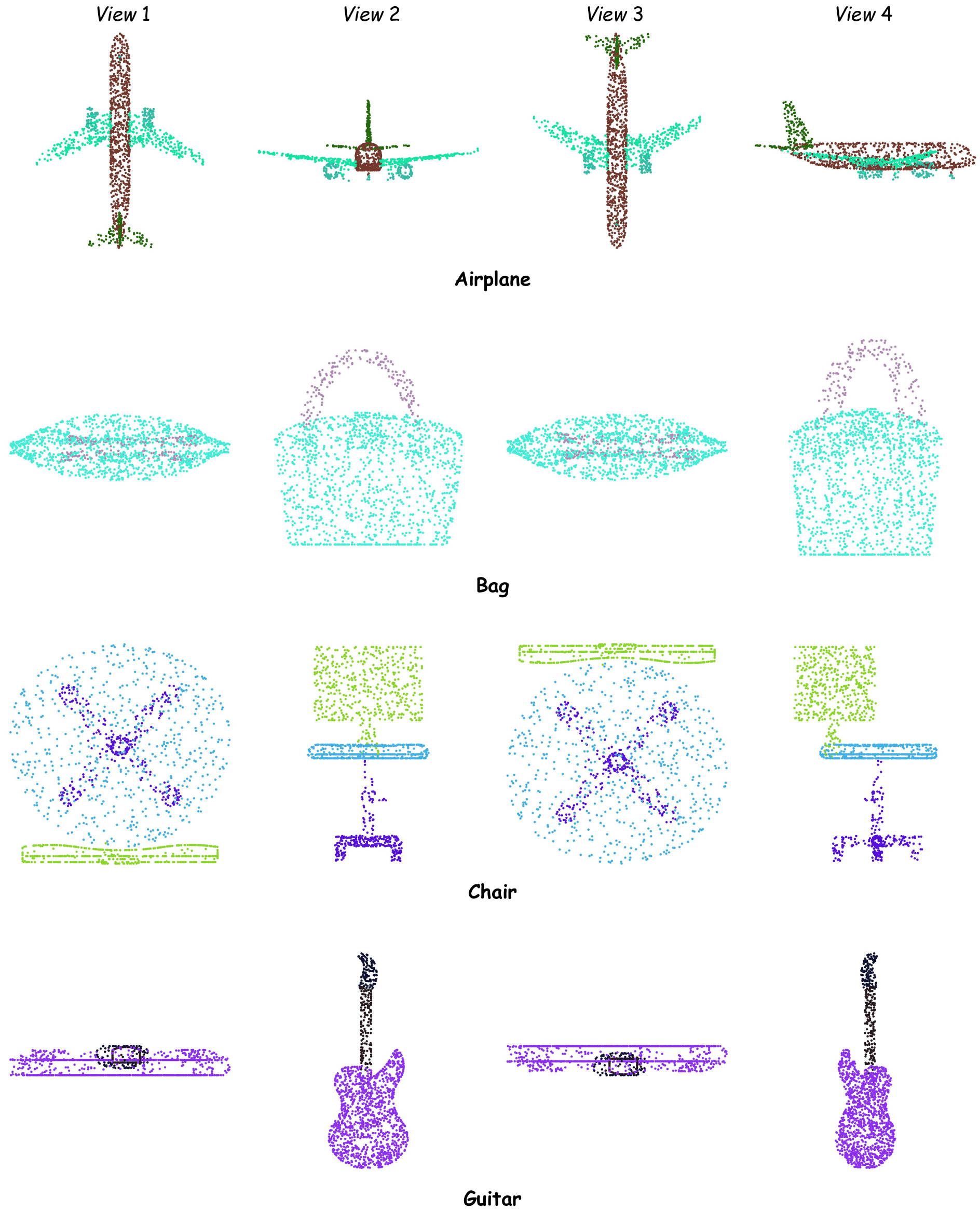}
    \caption{Visualization results for part segmentation on ShapeNetPart~\cite{yi2016scalable}. We present projected prediction images from \textbf{\texttt{\textcolor{lora_blue}{Point}\textcolor{lora_orange}{LoRA}}} across four different viewpoints, including ``Airplane", ``Bag", ``Chair" and ``Guitar".}
    \label{fig:supp_fig1}
\end{figure*}

\begin{figure*}[t]
    \centering
    \includegraphics[width=\linewidth]{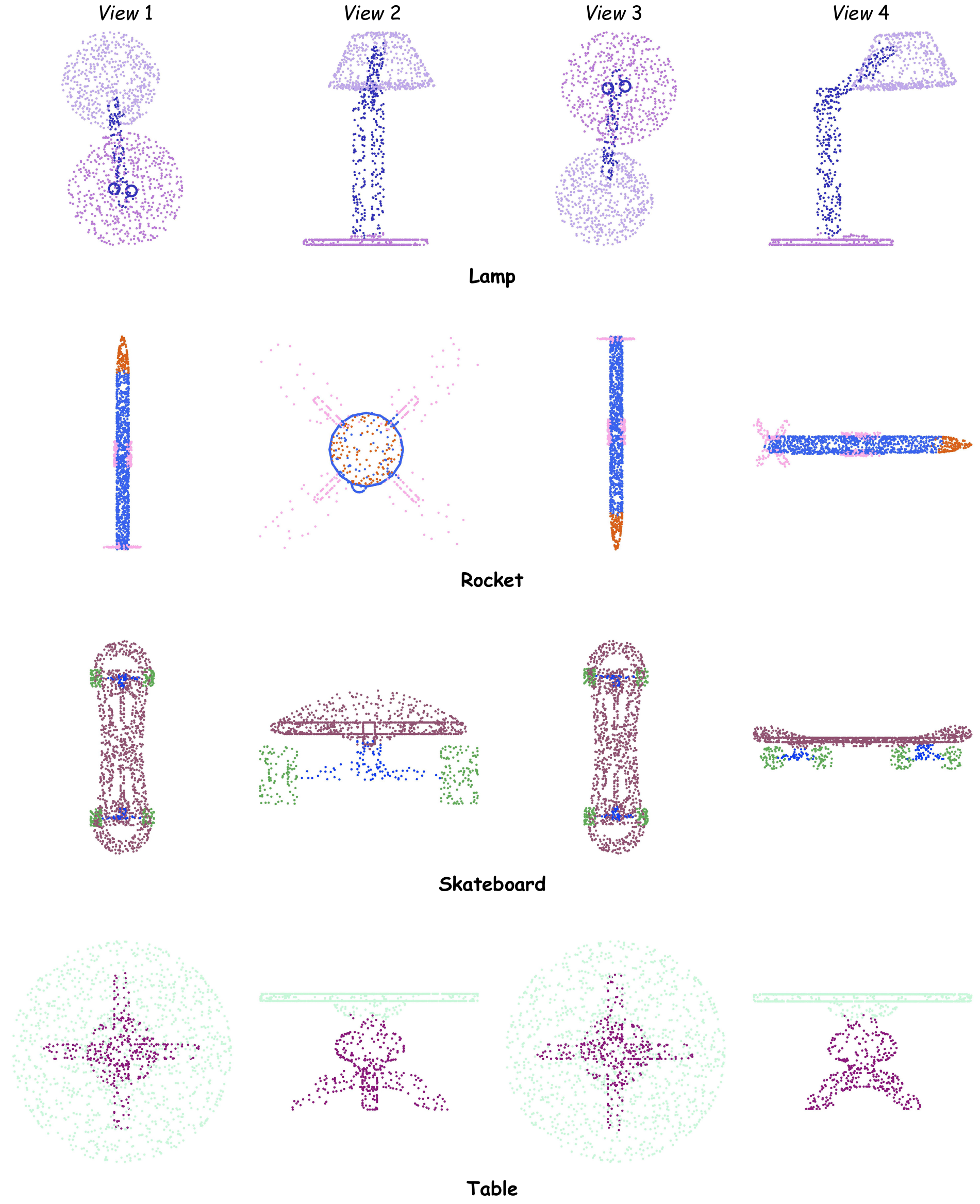}
    \caption{Visualization results for part segmentation on ShapeNetPart~\cite{yi2016scalable}. Projected prediction images from \textbf{\texttt{\textcolor{lora_blue}{Point}\textcolor{lora_orange}{LoRA}}} are shown across four different viewpoints, including the categories ``Lamp", ``Rocket", ``Skateboard" and ``Table".}
    \label{fig:supp_fig2}
\end{figure*}

We visualize the results of part segmentation obtained using the proposed approach, fine-tuned with the Recon~\cite{qi2023recon} pretrained model in ShapeNetPart~\cite{yi2016scalable}. 
As illustrated in \cref{fig:supp_fig1} and \cref{fig:supp_fig2}, a total of eight representative categories are selected, with four viewpoints displayed for each category. 
Our method demonstrates promising segmentation performance across various categories while utilizing a minimal number of tunable parameters.


\section{Additional Discussions}
\label{sec:supp_c}
\subsection{Explanatory Experiments and Discussions}
\input{tables/supp_large_model}
\noindent \textbf{Large model experiments and necessity for PEFT.} The experiments in the main paper follow the \textit{common settings}, validating our approach on small-scale models ($22.1$M) for fair comparison. 
This establishes a solid foundation for the extension to larger-scale models. 
We further fine-tune PointGPT-L~\cite{chen2024pointgpt, liang2024parameter} ($360.5$M), the largest pre-trained model for object-level point clouds, using proposed PointLoRA on PB-T50-RS, 
As shown in Tab.~\ref{tab:supp_large}, our method updates only $1.36\%$ of parameters and outperforms full fine-tuning with significantly reduced storage space.

\noindent \textbf{About the technical novelty of PointLoRA.} 
First, we reveal the effectiveness of LoRA in point cloud and its connection to PointNet, which is \textit{overlooked in previous research}. 
Second, adhering to the principle of \textit{simplicity and effectiveness}, we design PointLoRA with multi-scale token selection that requires only minimal parameters to achieve SOTA performance. This simplicity and efficiency enable seamless extension to larger models and diverse scenarios.

\noindent \textbf{Theoretical analysis of LoRA for point cloud.} 
LoRA is well-suited for point clouds due to its alignment with the principles underlying point cloud architectures like PointNet. Both leverage efficient subspace representations: PointNet adopts \textit{shared MLPs and pooling} to approximate \textit{permutation-invariant} set functions, while LoRA reduces fine-tuning updates with \textit{low-rank matrices}. This synergy allows LoRA to effectively adapt to the sparse, high-dimensional nature of point clouds to capture \textit{ global features} with minimal computational overhead.

\subsection{Limitations and Future Work}
While \textbf{\texttt{\textcolor{lora_blue}{Point}\textcolor{lora_orange}{LoRA}}} effectively reduces trainable parameters and achieves competitive performance across diverse tasks, it still has certain limitations. 
The effectiveness of fine-tuning heavily depends on the quality of pre-trained models, making it less adaptable to tasks involving domains significantly different from those used during pre-training. 
Additionally, the multi-scale token selection strategy is heuristically designed, and its performance may vary across various datasets and tasks. 
Furthermore, the scalability of our method to extremely large pre-trained models remains unexamined, partly due to the current absence of general large-scale models in 3D space.
The variation in task-specific performance also highlights the need for more tailored solutions.

Future research could focus on developing more adaptive or learnable token selection mechanisms to enhance flexibility and robustness. 
Exploring task-conditioned fine-tuning strategies and hierarchical LoRA configurations may improve scalability and performance, particularly for larger models. 
Expanding the approach to handle multi-modal data, such as combining point clouds with images or text, presents another promising direction. 
Meanwhile, investigating domain-specific adaptation techniques could improve performance in scenarios with significant domain shifts from pre-training to downstream tasks.

\subsection{Broader Impacts} 
The proposed approach facilitates parameter-efficient fine-tuning for pre-trained point cloud models, increasing accessibility to advanced technologies in domains such as autonomous driving, robotics, and environmental monitoring. 
Its efficiency also contributes to reducing the environmental impact of deep learning by lowering energy consumption. 
However, the improved capabilities of point cloud modeling present risks, including potential misuse in privacy-invasive applications or the propagation of unintended biases in autonomous systems. To maximize its benefits while addressing these challenges, ethical deployment and responsible governance will be essential.

\section{License and Consent Information}

\label{sec:supp_d}
\subsection{Public Datasets}
We conducted all the experiments on the subsequent openly accessible datasets:
\begin{itemize}
    \item ScanObjectNN~\cite{uy2019revisiting}\footnote{\url{https://hkust-vgd.github.io/scanobjectnn}.} \dotfill MIT License
    \item ModelNet40~\cite{wu20153d}\footnote{\url{https://modelnet.cs.princeton.edu}.} \dotfill Other (specified in description)
    \item  ShapeNetPart~\cite{yi2016scalable}\footnote{\url{https://cs.stanford.edu/~ericyi/project_page/part_annotation}.} \dotfill Other (specified in description)
\end{itemize}

\subsection{Public Implementation}
We compare and validate the effectiveness of the proposed method with the following publicly available pre-trained models and source codes:
\begin{itemize}
     \item Point-MAE~\cite{pang2022masked}\footnote{\url{https://github.com/Pang-Yatian/Point-MAE}.} \dotfill MIT License
     \item ReCon~\cite{qi2023recon}\footnote{\url{https://github.com/qizekun/ReCon}.} \dotfill MIT License
     \item Point-BERT~\cite{yu2022point}\footnote{\url{https://github.com/Julie-tang00/Point-BERT}.} \dotfill MIT License
     \item  IDPT~\cite{zha2023instance}\footnote{\url{https://github.com/zyh16143998882/ICCV23-IDPT}.} \dotfill Other (specified in description)
     \item DAPT~\cite{zhou2024dynamic}\footnote{\url{https://github.com/LMD0311/DAPT}.} \dotfill Apache License 2.0
     \item PPT~\cite{zhang2024positional}\footnote{\url{https://github.com/zsc000722/PPT}.} \dotfill MIT License
     \item LoRA~\cite{hu2021lora}\footnote{\url{https://github.com/microsoft/LoRA}.} \dotfill MIT License
    
\end{itemize}

%% file: tables/supp_abl_msfps.tex
\begin{table}[h]
    \centering
    \caption{
        Ablation study on the number of center points and neighbor points in multi-scale token selection.
    }
    \vspace{-0.2cm}
    \footnotesize
    \renewcommand\tabcolsep{15pt}
    \renewcommand\arraystretch{1.1}
    \begin{tabular}{ccc}
        \toprule
         \textbf{Scale $1$} & \textbf{Scale $2$} & \textbf{PB-T50-RS} \\
        \midrule\midrule
          $(256, 16)$ & $(64, 64)$ &  $84.46$  \\
          $(256, 16)$ & $(128, 32)$ & $84.66$  \\
          $(128, 16)$ & $(64, 32)$ &  $85.08$ \\
          $(128, 48)$ & $(64, 80)$ & $83.90$  \\
         \rowcolor{lora_orange!10} $(128,32)$ & $(64,64)$ & $\mathbf{85.53}$  \\
        \bottomrule
    \end{tabular}
    \label{tab:abl_msfps}
    \vspace{-2mm}
\end{table}

%% file: tables/supp_abl_loss.tex
\begin{table}[h]
    \footnotesize
    \centering
    \renewcommand\tabcolsep{4.5pt}
    \renewcommand\arraystretch{1.1}
    \caption{Ablation on the loss weight for Mask Predictor learning. }
    \vspace{-0.2cm}
    \begin{tabular}[b]{cccccc}
        \toprule
          \textbf{Weight ($\lambda$)}  & $0$ & $0.002$ & \cellcolor{lora_orange!10}$0.004$ & $0.006$ & $0.008$   \\
        \midrule\midrule
         \textbf{PB-T50-RS}  &  $84.56$ & $84.90$ &  \cellcolor{lora_orange!10}$\mathbf{85.53}$  & $84.91$ & $85.01$ \\
        \bottomrule
    \end{tabular}
    \vspace{-2mm}
    \label{tab:abl_loss}
\end{table}

%% file: tables/supp_abl_layer.tex
\begin{table}[h]
    \renewcommand\tabcolsep{10pt}
    \renewcommand\arraystretch{1.1}
        \centering
         \footnotesize
     \caption{Ablation study on the injected blocks for \textbf{\texttt{\textcolor{lora_blue}{Point}\textcolor{lora_orange}{LoRA}}}.}
     \vspace{-0.2cm}
        \begin{tabular}{cccc}
            \toprule
            \textbf{Blocks} &  \textbf{TP} & \textbf{Ratio} & \textbf{PB\_T50\_RS} \\
            \midrule\midrule
            $1\rightarrow3$  & $0.61$ M & $2.72\%$ & $83.83$ \\
            $1\rightarrow6$  & $0.66$ M & $2.96\%$ & $83.55$ \\
            $1\rightarrow9$  & $0.72$ M & $3.19\%$ & $85.05$\\
            $4\rightarrow12$ & $0.72$ M & $3.19\%$ & $84.84$\\
            $8\rightarrow12$ & $0.64$ M & $2.88\%$ & $84.14$ \\
            \rowcolor{lora_orange!10} $1\rightarrow12$  & $0.77$ M & $3.43\%$ & {$\mathbf{85.53}$}\\
            \bottomrule 
        \end{tabular}
    \label{tab:insert}
    \vspace{-0.2cm}
\end{table}

%% file: tables/supp_training.tex
\begin{table*}[t]
\footnotesize
\centering
\setlength{\tabcolsep}{5.8mm}
\caption{Training settings for various downstream fine-tuning models and datasets used in our implementation.}
\vspace{-0.2cm}
\begin{tabular}{lccccc}
\toprule
    \multirow{2.3}{*}{\textbf{Training Settings}}  &\multicolumn{3}{c}{\textbf{Classification}} & \multicolumn{2}{c}{\textbf{Segmentation}}\\
		\cmidrule(r){2-4} \cmidrule(r){5-6}
	 &ScanObjectNN~\cite{uy2019revisiting} & ModelNet40~\cite{wu20153d} & ModelNet40 Few-shot~\cite{wu20153d} & ShapeNetPart~\cite{yi2016scalable}     \\
    \midrule\midrule
 Pre-trained Model & Point-MAE~\cite{pang2022masked} & Point-MAE~\cite{pang2022masked} & Recon~\cite{qi2023recon} & Recon~\cite{qi2023recon} \\
 Optimizer & AdamW & AdamW & AdamW & AdamW \\
 Learning rate & $5\times10^{-4}$ & $5\times10^{-4}$ & $5\times10^{-4}$  & $2\times10^{-4}$\\
 Weight decay & $5\times10^{-2}$ & $5\times10^{-2}$ & $5\times10^{-2}$ & $5\times10^{-2}$ \\
 Learning rate scheduler & cosine & cosine & cosine & cosine \\
 Training epochs  & $300$ & $300$ & $150$ & $300$ \\
 Warm-up epochs& $10$ & $10$ & $10$ & $10$ \\
 Batch size & $32$ & $32$ & $4$ & $16$ \\
 Drop path rate & $0.3$ & $0.1$ & $0.3$ & $0.1$ \\
Selected token number & $32$ \& $8$   & $32$ \& $8$ & $32$ \& $8$ & $32$ \& $8$  \\
 \midrule
 Number of points  & $2048$ & $1024$ & $1024$ & $2048$ \\
 Number of point patches & $128$ & $64$ & $64$ & $128$\\
 Point patch size  & $32$ & $32$ & $32$  & $32$ \\
\bottomrule
\end{tabular}
\label{tab:training}
\end{table*}

%% file: tables/supp_large_model.tex
\begin{table}[t]
    \footnotesize
    \centering
    \renewcommand\tabcolsep{2.8pt}
    \renewcommand\arraystretch{1.1}
    \caption{Explanatory experiments on large model with the proposed method.}
    \vspace{-3.5mm}
    \scalebox{1.0}{
    \begin{tabular}[b]{c|ccc}
        \toprule
          \textbf{Methods}  & \textbf{Tunable Params.}  & \textbf{Storage} & \textbf{PB-T50-RS} \\
        \midrule\midrule
         {\textbf{PointGPT-L} (Full-FT)} & {$360.5$ M} & {$4.0$ GB} & {$93.4$} \\
          \rowcolor{lora_orange!10} +\textbf{\texttt{\textcolor{lora_blue}{Point}\textcolor{lora_orange}{LoRA}}} & $\mathbf{4.9}$ \textbf{M}  &  $\mathbf{<60}$ \textbf{MB}  & $\mathbf{93.8}$ \\
        \bottomrule
    \end{tabular}}
    \label{tab:supp_large}
    \vspace{-4.5mm}
\end{table}

%% file: main.bbl
\begin{thebibliography}{77}
\providecommand{\natexlab}[1]{#1}
\providecommand{\url}[1]{\texttt{#1}}
\expandafter\ifx\csname urlstyle\endcsname\relax
  \providecommand{\doi}[1]{doi: #1}\else
  \providecommand{\doi}{doi: \begingroup \urlstyle{rm}\Url}\fi

\bibitem[Afham et~al.(2022)Afham, Dissanayake, Dissanayake, Dharmasiri, Thilakarathna, and Rodrigo]{afham2022crosspoint}
Mohamed Afham, Isuru Dissanayake, Dinithi Dissanayake, Amaya Dharmasiri, Kanchana Thilakarathna, and Ranga Rodrigo.
\newblock Crosspoint: Self-supervised cross-modal contrastive learning for 3d point cloud understanding.
\newblock In \emph{IEEE/CVF Conference on Computer Vision and Pattern Recognition}, pages 9902--9912, 2022.

\bibitem[Ba(2016)]{ba2016layer}
Jimmy~Lei Ba.
\newblock Layer normalization.
\newblock \emph{arXiv preprint arXiv:1607.06450}, 2016.

\bibitem[Bebis and Georgiopoulos(1994)]{bebis1994feed}
George Bebis and Michael Georgiopoulos.
\newblock Feed-forward neural networks.
\newblock \emph{Ieee Potentials}, 13\penalty0 (4):\penalty0 27--31, 1994.

\bibitem[Bello et~al.(2020)Bello, Yu, Wang, Adam, and Li]{bello2020deep}
Saifullahi~Aminu Bello, Shangshu Yu, Cheng Wang, Jibril~Muhmmad Adam, and Jonathan Li.
\newblock Deep learning on 3d point clouds.
\newblock \emph{Remote Sensing}, 12\penalty0 (11):\penalty0 1729, 2020.

\bibitem[Chen et~al.(2024)Chen, Wang, Yang, Yu, Yuan, and Yue]{chen2024pointgpt}
Guangyan Chen, Meiling Wang, Yi Yang, Kai Yu, Li Yuan, and Yufeng Yue.
\newblock Pointgpt: Auto-regressively generative pre-training from point clouds.
\newblock In \emph{Advances in Neural Information Processing Systems}, 2024.

\bibitem[Chen et~al.(2023{\natexlab{a}})Chen, Gu, Liu, Magid, Dong, Wang, Pfister, and Zhu]{chen2023masked}
Haoyu Chen, Jinjin Gu, Yihao Liu, Salma~Abdel Magid, Chao Dong, Qiong Wang, Hanspeter Pfister, and Lei Zhu.
\newblock Masked image training for generalizable deep image denoising.
\newblock In \emph{IEEE/CVF Conference on Computer Vision and Pattern Recognition}, pages 1692--1703, 2023{\natexlab{a}}.

\bibitem[Chen et~al.(2023{\natexlab{b}})Chen, Liu, Kong, Zhu, Ma, Li, Hou, Qiao, and Wang]{chen2023clip2scene}
Runnan Chen, Youquan Liu, Lingdong Kong, Xinge Zhu, Yuexin Ma, Yikang Li, Yuenan Hou, Yu Qiao, and Wenping Wang.
\newblock Clip2scene: Towards label-efficient 3d scene understanding by clip.
\newblock In \emph{IEEE/CVF Conference on Computer Vision and Pattern Recognition}, pages 7020--7030, 2023{\natexlab{b}}.

\bibitem[Chen et~al.(2022)Chen, Ge, Tong, Wang, Song, Wang, and Luo]{chen2022adaptformer}
Shoufa Chen, Chongjian Ge, Zhan Tong, Jiangliu Wang, Yibing Song, Jue Wang, and Ping Luo.
\newblock Adaptformer: Adapting vision transformers for scalable visual recognition.
\newblock In \emph{Advances in Neural Information Processing Systems}, 2022.

\bibitem[Ding et~al.(2023)Ding, Qin, Yang, Wei, Yang, Su, Hu, Chen, Chan, Chen, et~al.]{ding2023parameter}
Ning Ding, Yujia Qin, Guang Yang, Fuchao Wei, Zonghan Yang, Yusheng Su, Shengding Hu, Yulin Chen, Chi-Min Chan, Weize Chen, et~al.
\newblock Parameter-efficient fine-tuning of large-scale pre-trained language models.
\newblock \emph{Nature Machine Intelligence}, 5\penalty0 (3):\penalty0 220--235, 2023.

\bibitem[Dong et~al.(2022)Dong, Qi, Zhang, Zhang, Sun, Ge, Yi, and Ma]{dong2022autoencoders}
Runpei Dong, Zekun Qi, Linfeng Zhang, Junbo Zhang, Jianjian Sun, Zheng Ge, Li Yi, and Kaisheng Ma.
\newblock Autoencoders as cross-modal teachers: Can pretrained 2d image transformers help 3d representation learning?
\newblock In \emph{International Conference on Learning Representations}, 2022.

\bibitem[Dosovitskiy et~al.(2021)Dosovitskiy, Beyer, Kolesnikov, Weissenborn, Zhai, Unterthiner, Dehghani, Minderer, Heigold, Gelly, et~al.]{dosovitskiy2020image}
Alexey Dosovitskiy, Lucas Beyer, Alexander Kolesnikov, Dirk Weissenborn, Xiaohua Zhai, Thomas Unterthiner, Mostafa Dehghani, Matthias Minderer, Georg Heigold, Sylvain Gelly, et~al.
\newblock An image is worth 16x16 words: Transformers for image recognition at scale.
\newblock In \emph{International Conference on Learning Representations}, 2021.

\bibitem[Guo et~al.(2020)Guo, Wang, Hu, Liu, Liu, and Bennamoun]{guo2020deep}
Yulan Guo, Hanyun Wang, Qingyong Hu, Hao Liu, Li Liu, and Mohammed Bennamoun.
\newblock Deep learning for 3d point clouds: A survey.
\newblock \emph{IEEE transactions on pattern analysis and machine intelligence}, 43\penalty0 (12):\penalty0 4338--4364, 2020.

\bibitem[Hamdi et~al.(2021)Hamdi, Giancola, and Ghanem]{hamdi2021mvtn}
Abdullah Hamdi, Silvio Giancola, and Bernard Ghanem.
\newblock Mvtn: Multi-view transformation network for 3d shape recognition.
\newblock In \emph{IEEE/CVF International Conference on Computer Vision}, pages 1--11, 2021.

\bibitem[He et~al.(2021)He, Zhou, Ma, Berg-Kirkpatrick, and Neubig]{he2021towards}
Junxian He, Chunting Zhou, Xuezhe Ma, Taylor Berg-Kirkpatrick, and Graham Neubig.
\newblock Towards a unified view of parameter-efficient transfer learning.
\newblock In \emph{International Conference on Learning Representations}, 2021.

\bibitem[He et~al.(2022)He, Chen, Xie, Li, Doll{\'a}r, and Girshick]{he2022masked}
Kaiming He, Xinlei Chen, Saining Xie, Yanghao Li, Piotr Doll{\'a}r, and Ross Girshick.
\newblock Masked autoencoders are scalable vision learners.
\newblock In \emph{IEEE/CVF Conference on Computer Vision and Pattern Recognition}, 2022.

\bibitem[Hendrycks and Gimpel(2016)]{hendrycks2016gaussian}
Dan Hendrycks and Kevin Gimpel.
\newblock Gaussian error linear units (gelus).
\newblock \emph{arXiv preprint arXiv:1606.08415}, 2016.

\bibitem[Hinton et~al.(2015)Hinton, Vinyals, and Dean]{hinton2015distilling}
Geoffrey Hinton, Oriol Vinyals, and Jeff Dean.
\newblock Distilling the knowledge in a neural network.
\newblock \emph{arXiv preprint arXiv:1503.02531}, 2015.

\bibitem[Hong et~al.(2023)Hong, Chou, and Liu]{hong2023attention}
Cheng-Yao Hong, Yu-Ying Chou, and Tyng-Luh Liu.
\newblock Attention discriminant sampling for point clouds.
\newblock In \emph{IEEE/CVF International Conference on Computer Vision}, pages 14429--14440, 2023.

\bibitem[Houlsby et~al.(2019)Houlsby, Giurgiu, Jastrzebski, Morrone, De~Laroussilhe, Gesmundo, Attariyan, and Gelly]{houlsby2019parameter}
Neil Houlsby, Andrei Giurgiu, Stanislaw Jastrzebski, Bruna Morrone, Quentin De~Laroussilhe, Andrea Gesmundo, Mona Attariyan, and Sylvain Gelly.
\newblock Parameter-efficient transfer learning for nlp.
\newblock In \emph{International Conference on Machine Learning}. PMLR, 2019.

\bibitem[Hu et~al.(2021)Hu, Wallis, Allen-Zhu, Li, Wang, Wang, Chen, et~al.]{hu2021lora}
Edward~J Hu, Phillip Wallis, Zeyuan Allen-Zhu, Yuanzhi Li, Shean Wang, Lu Wang, Weizhu Chen, et~al.
\newblock Lora: Low-rank adaptation of large language models.
\newblock In \emph{International Conference on Learning Representations}, 2021.

\bibitem[Jia et~al.(2022)Jia, Tang, Chen, Cardie, Belongie, Hariharan, and Lim]{jia2022visual}
Menglin Jia, Luming Tang, Bor-Chun Chen, Claire Cardie, Serge Belongie, Bharath Hariharan, and Ser-Nam Lim.
\newblock Visual prompt tuning.
\newblock In \emph{European Conference on Computer Vision}, 2022.

\bibitem[Kong et~al.(2023{\natexlab{a}})Kong, Liu, Chen, Ma, Zhu, Li, Hou, Qiao, and Liu]{kong2023rethinking}
Lingdong Kong, Youquan Liu, Runnan Chen, Yuexin Ma, Xinge Zhu, Yikang Li, Yuenan Hou, Yu Qiao, and Ziwei Liu.
\newblock Rethinking range view representation for lidar segmentation.
\newblock In \emph{IEEE/CVF International Conference on Computer Vision}, pages 228--240, 2023{\natexlab{a}}.

\bibitem[Kong et~al.(2023{\natexlab{b}})Kong, Liu, Li, Chen, Zhang, Ren, Pan, Chen, and Liu]{kong2023robo3d}
Lingdong Kong, Youquan Liu, Xin Li, Runnan Chen, Wenwei Zhang, Jiawei Ren, Liang Pan, Kai Chen, and Ziwei Liu.
\newblock Robo3d: Towards robust and reliable 3d perception against corruptions.
\newblock In \emph{IEEE/CVF International Conference on Computer Vision}, pages 19994--20006, 2023{\natexlab{b}}.

\bibitem[Kong et~al.(2023{\natexlab{c}})Kong, Ren, Pan, and Liu]{kong2023lasermix}
Lingdong Kong, Jiawei Ren, Liang Pan, and Ziwei Liu.
\newblock Lasermix for semi-supervised lidar semantic segmentation.
\newblock In \emph{IEEE/CVF Conference on Computer Vision and Pattern Recognition}, pages 21705--21715, 2023{\natexlab{c}}.

\bibitem[Lester et~al.(2021)Lester, Al-Rfou, and Constant]{lester2021power}
Brian Lester, Rami Al-Rfou, and Noah Constant.
\newblock The power of scale for parameter-efficient prompt tuning.
\newblock In \emph{Empirical Methods in Natural Language Processing}, pages 3045--3059, 2021.

\bibitem[Li et~al.(2023)Li, Hu, Nie, Han, Jiang, Guo, and Liu]{li2023dropkey}
Bonan Li, Yinhan Hu, Xuecheng Nie, Congying Han, Xiangjian Jiang, Tiande Guo, and Luoqi Liu.
\newblock Dropkey for vision transformer.
\newblock In \emph{IEEE/CVF Conference on Computer Vision and Pattern Recognition}, pages 22700--22709, 2023.

\bibitem[Li et~al.(2024{\natexlab{a}})Li, Ye, Huang, Zhang, Chen, He, Fan, and Ouyang]{li2024adapter}
Minglei Li, Peng Ye, Yongqi Huang, Lin Zhang, Tao Chen, Tong He, Jiayuan Fan, and Wanli Ouyang.
\newblock Adapter-x: A novel general parameter-efficient fine-tuning framework for vision.
\newblock \emph{arXiv preprint arXiv:2406.03051}, 2024{\natexlab{a}}.

\bibitem[Li and Liang(2021)]{li2021prefix}
Xiang~Lisa Li and Percy Liang.
\newblock Prefix-tuning: Optimizing continuous prompts for generation.
\newblock In \emph{Annual Meeting of the Association for Computational Linguistics}, 2021.

\bibitem[Li et~al.(2018)Li, Bu, Sun, Wu, Di, and Chen]{li2018pointcnn}
Yangyan Li, Rui Bu, Mingchao Sun, Wei Wu, Xinhan Di, and Baoquan Chen.
\newblock Pointcnn: Convolution on x-transformed points.
\newblock In \emph{Advances in Neural Information Processing Systems}, 2018.

\bibitem[Li et~al.(2024{\natexlab{b}})Li, Kong, Hu, Xu, and Huang]{li2024is}
Ye Li, Lingdong Kong, Hanjiang Hu, Xiaohao Xu, and Xiaonan Huang.
\newblock Is your lidar placement optimized for 3d scene understanding?
\newblock In \emph{Advances in Neural Information Processing Systems}, pages 34980--35017, 2024{\natexlab{b}}.

\bibitem[Lian et~al.(2022)Lian, Zhou, Feng, and Wang]{lian2022scaling}
Dongze Lian, Daquan Zhou, Jiashi Feng, and Xinchao Wang.
\newblock Scaling \& shifting your features: A new baseline for efficient model tuning.
\newblock In \emph{Advances in Neural Information Processing Systems}, 2022.

\bibitem[Liang et~al.(2024{\natexlab{a}})Liang, Feng, Zhou, Zhang, Zou, and Bai]{liang2024parameter}
Dingkang Liang, Tianrui Feng, Xin Zhou, Yumeng Zhang, Zhikang Zou, and Xiang Bai.
\newblock Parameter-efficient fine-tuning in spectral domain for point cloud learning.
\newblock \emph{arXiv preprint arXiv:2410.08114}, 2024{\natexlab{a}}.

\bibitem[Liang et~al.(2024{\natexlab{b}})Liang, Zhou, Wang, Zhu, Xu, Zou, Ye, and Bai]{liang2024pointmamba}
Dingkang Liang, Xin Zhou, Xinyu Wang, Xingkui Zhu, Wei Xu, Zhikang Zou, Xiaoqing Ye, and Xiang Bai.
\newblock Pointmamba: A simple state space model for point cloud analysis.
\newblock \emph{arXiv preprint arXiv:2402.10739}, 2024{\natexlab{b}}.

\bibitem[Liu et~al.(2022)Liu, Cai, and Lee]{liu2022masked}
Haotian Liu, Mu Cai, and Yong~Jae Lee.
\newblock Masked discrimination for self-supervised learning on point clouds.
\newblock In \emph{European Conference on Computer Vision}, 2022.

\bibitem[Liu et~al.(2023{\natexlab{a}})Liu, Wu, Zhao, Zhu, Xu, Tian, and Zheng]{liu2023moelora}
Qidong Liu, Xian Wu, Xiangyu Zhao, Yuanshao Zhu, Derong Xu, Feng Tian, and Yefeng Zheng.
\newblock Moelora: An moe-based parameter efficient fine-tuning method for multi-task medical applications.
\newblock \emph{arXiv preprint arXiv:2310.18339}, 2023{\natexlab{a}}.

\bibitem[Liu et~al.(2023{\natexlab{b}})Liu, Kong, Cen, Chen, Zhang, Pan, Chen, and Liu]{liu2023seal}
Youquan Liu, Lingdong Kong, Jun Cen, Runnan Chen, Wenwei Zhang, Liang Pan, Kai Chen, and Ziwei Liu.
\newblock Segment any point cloud sequences by distilling vision foundation models.
\newblock In \emph{Advances in Neural Information Processing Systems}, pages 37193--37229, 2023{\natexlab{b}}.

\bibitem[Ma et~al.(2022)Ma, Qin, You, Ran, and Fu]{ma2022rethinking}
Xu Ma, Can Qin, Haoxuan You, Haoxi Ran, and Yun Fu.
\newblock Rethinking network design and local geometry in point cloud: A simple residual mlp framework.
\newblock In \emph{International Conference on Learning Representations}, 2022.

\bibitem[Mann et~al.(2020)Mann, Ryder, Subbiah, Kaplan, Dhariwal, Neelakantan, Shyam, Sastry, Askell, Agarwal, et~al.]{mann2020language}
Ben Mann, N Ryder, M Subbiah, J Kaplan, P Dhariwal, A Neelakantan, P Shyam, G Sastry, A Askell, S Agarwal, et~al.
\newblock Language models are few-shot learners.
\newblock \emph{arXiv preprint arXiv:2005.14165}, 1, 2020.

\bibitem[Nunes et~al.(2022)Nunes, Marcuzzi, Chen, Behley, and Stachniss]{nunes2022segcontrast}
Lucas Nunes, Rodrigo Marcuzzi, Xieyuanli Chen, Jens Behley, and Cyrill Stachniss.
\newblock Segcontrast: 3d point cloud feature representation learning through self-supervised segment discrimination.
\newblock \emph{IEEE Robotics and Automation Letters}, 7\penalty0 (2):\penalty0 2116--2123, 2022.

\bibitem[Pang et~al.(2022)Pang, Wang, Tay, Liu, Tian, and Yuan]{pang2022masked}
Yatian Pang, Wenxiao Wang, Francis~EH Tay, Wei Liu, Yonghong Tian, and Li Yuan.
\newblock Masked autoencoders for point cloud self-supervised learning.
\newblock In \emph{European Conference on Computer Vision}, 2022.

\bibitem[Qi et~al.(2017{\natexlab{a}})Qi, Su, Mo, and Guibas]{qi2017pointnet}
Charles~R Qi, Hao Su, Kaichun Mo, and Leonidas~J Guibas.
\newblock Pointnet: Deep learning on point sets for 3d classification and segmentation.
\newblock In \emph{IEEE/CVF Conference on Computer Vision and Pattern Recognition}, 2017{\natexlab{a}}.

\bibitem[Qi et~al.(2017{\natexlab{b}})Qi, Yi, Su, and Guibas]{qi2017pointnet++}
Charles~Ruizhongtai Qi, Li Yi, Hao Su, and Leonidas~J Guibas.
\newblock Pointnet++: Deep hierarchical feature learning on point sets in a metric space.
\newblock In \emph{Advances in Neural Information Processing Systems}, 2017{\natexlab{b}}.

\bibitem[Qi et~al.(2023)Qi, Dong, Fan, Ge, Zhang, Ma, and Yi]{qi2023recon}
Zekun Qi, Runpei Dong, Guofan Fan, Zheng Ge, Xiangyu Zhang, Kaisheng Ma, and Li Yi.
\newblock Contrast with reconstruct: Contrastive 3d representation learning guided by generative pretraining.
\newblock In \emph{International Conference on Machine Learning}. PMLR, 2023.

\bibitem[Qi et~al.(2024)Qi, Dong, Zhang, Geng, Han, Ge, Yi, and Ma]{qi2024shapellm}
Zekun Qi, Runpei Dong, Shaochen Zhang, Haoran Geng, Chunrui Han, Zheng Ge, Li Yi, and Kaisheng Ma.
\newblock Shapellm: Universal 3d object understanding for embodied interaction.
\newblock \emph{arXiv preprint arXiv:2402.17766}, 2024.

\bibitem[Qian et~al.(2022)Qian, Li, Peng, Mai, Hammoud, Elhoseiny, and Ghanem]{qian2022pointnext}
Guocheng Qian, Yuchen Li, Houwen Peng, Jinjie Mai, Hasan Hammoud, Mohamed Elhoseiny, and Bernard Ghanem.
\newblock Pointnext: Revisiting pointnet++ with improved training and scaling strategies.
\newblock In \emph{Advances in Neural Information Processing Systems}, pages 23192--23204, 2022.

\bibitem[Ran et~al.(2022)Ran, Liu, and Wang]{ran2022surface}
Haoxi Ran, Jun Liu, and Chengjie Wang.
\newblock Surface representation for point clouds.
\newblock In \emph{IEEE/CVF Conference on Computer Vision and Pattern Recognition}, 2022.

\bibitem[Sautier et~al.(2022)Sautier, Puy, Gidaris, Boulch, Bursuc, and Marlet]{sautier2022slidr}
Corentin Sautier, Gilles Puy, Spyros Gidaris, Alexandre Boulch, Andrei Bursuc, and Renaud Marlet.
\newblock Image-to-lidar self-supervised distillation for autonomous driving data.
\newblock In \emph{IEEE/CVF Conference on Computer Vision and Pattern Recognition}, pages 9891--9901, 2022.

\bibitem[Shi and Lipani(2024)]{shi2024dept}
Zhengxiang Shi and Aldo Lipani.
\newblock Dept: Decomposed prompt tuning for parameter-efficient fine-tuning.
\newblock In \emph{International Conference on Learning Representations}, 2024.

\bibitem[Song et~al.(2022)Song, Yu, Chen, and Yang]{song2022transformer}
Zikai Song, Junqing Yu, Yi-Ping~Phoebe Chen, and Wei Yang.
\newblock Transformer tracking with cyclic shifting window attention.
\newblock In \emph{IEEE/CVF Conference on Computer Vision and Pattern Recognition}, pages 8791--8800, 2022.

\bibitem[Sung et~al.(2022)Sung, Cho, and Bansal]{sung2022lst}
Yi-Lin Sung, Jaemin Cho, and Mohit Bansal.
\newblock Lst: Ladder side-tuning for parameter and memory efficient transfer learning.
\newblock In \emph{Advances in Neural Information Processing Systems}, pages 12991--13005, 2022.

\bibitem[Tang et~al.(2024)Tang, Zhang, Guo, Ma, Zhao, Wang, Wang, and Li]{tang2024point}
Yiwen Tang, Ray Zhang, Zoey Guo, Xianzheng Ma, Bin Zhao, Zhigang Wang, Dong Wang, and Xuelong Li.
\newblock Point-peft: Parameter-efficient fine-tuning for 3d pre-trained models.
\newblock In \emph{AAAI Conference on Artificial Intelligence}, pages 5171--5179, 2024.

\bibitem[Uy et~al.(2019)Uy, Pham, Hua, Nguyen, and Yeung]{uy2019revisiting}
Mikaela~Angelina Uy, Quang-Hieu Pham, Binh-Son Hua, Thanh Nguyen, and Sai-Kit Yeung.
\newblock Revisiting point cloud classification: A new benchmark dataset and classification model on real-world data.
\newblock In \emph{IEEE/CVF International Conference on Computer Vision}, 2019.

\bibitem[Van~der Maaten and Hinton(2008)]{van2008visualizing}
Laurens Van~der Maaten and Geoffrey Hinton.
\newblock Visualizing data using t-sne.
\newblock \emph{Journal of Machine Learning Research}, 2008.

\bibitem[Wang et~al.(2021)Wang, Liu, Yue, Lasenby, and Kusner]{wang2021unsupervised}
Hanchen Wang, Qi Liu, Xiangyu Yue, Joan Lasenby, and Matt~J Kusner.
\newblock Unsupervised point cloud pre-training via occlusion completion.
\newblock In \emph{IEEE/CVF International Conference on Computer Vision}, 2021.

\bibitem[Wang et~al.(2022)Wang, Zhu, and Zhang]{wang2022meta}
Song Wang, Jianke Zhu, and Ruixiang Zhang.
\newblock Meta-rangeseg: Lidar sequence semantic segmentation using multiple feature aggregation.
\newblock \emph{IEEE Robotics and Automation Letters}, 7\penalty0 (4):\penalty0 9739--9746, 2022.

\bibitem[Wang et~al.(2023)Wang, Li, Liu, Liu, and Zhu]{wang2023lidar2map}
Song Wang, Wentong Li, Wenyu Liu, Xiaolu Liu, and Jianke Zhu.
\newblock Lidar2map: In defense of lidar-based semantic map construction using online camera distillation.
\newblock In \emph{IEEE/CVF Conference on Computer Vision and Pattern Recognition}, pages 5186--5195, 2023.

\bibitem[Wang et~al.(2024{\natexlab{a}})Wang, Yu, Li, Liu, Liu, Chen, and Zhu]{wang2024not}
Song Wang, Jiawei Yu, Wentong Li, Wenyu Liu, Xiaolu Liu, Junbo Chen, and Jianke Zhu.
\newblock Not all voxels are equal: Hardness-aware semantic scene completion with self-distillation.
\newblock In \emph{IEEE/CVF Conference on Computer Vision and Pattern Recognition}, pages 14792--14801, 2024{\natexlab{a}}.

\bibitem[Wang et~al.(2024{\natexlab{b}})Wang, Yu, Li, Shi, Yang, Chen, and Zhu]{wang2024label}
Song Wang, Jiawei Yu, Wentong Li, Hao Shi, Kailun Yang, Junbo Chen, and Jianke Zhu.
\newblock Label-efficient semantic scene completion with scribble annotations.
\newblock In \emph{International Joint Conference on Artificial Intelligence}, pages 1398--1406, 2024{\natexlab{b}}.

\bibitem[Wang et~al.(2019)Wang, Sun, Liu, Sarma, Bronstein, and Solomon]{wang2019dynamic}
Yue Wang, Yongbin Sun, Ziwei Liu, Sanjay~E Sarma, Michael~M Bronstein, and Justin~M Solomon.
\newblock Dynamic graph cnn for learning on point clouds.
\newblock \emph{ACM Transactions on Graphics}, 38:\penalty0 1--12, 2019.

\bibitem[Wu et~al.(2023)Wu, Zheng, Pfrommer, and Beyerer]{wu2023attention}
Chengzhi Wu, Junwei Zheng, Julius Pfrommer, and J{\"u}rgen Beyerer.
\newblock Attention-based point cloud edge sampling.
\newblock In \emph{IEEE/CVF Conference on Computer Vision and Pattern Recognition}, 2023.

\bibitem[Wu et~al.(2015)Wu, Song, Khosla, Yu, Zhang, Tang, and Xiao]{wu20153d}
Zhirong Wu, Shuran Song, Aditya Khosla, Fisher Yu, Linguang Zhang, Xiaoou Tang, and Jianxiong Xiao.
\newblock 3d shapenets: A deep representation for volumetric shapes.
\newblock In \emph{IEEE/CVF Conference on Computer Vision and Pattern Recognition}, pages 1912--1920, 2015.

\bibitem[Xiao et~al.(2024{\natexlab{a}})Xiao, Zhang, Shao, and Lu]{xiao2024survey}
Aoran Xiao, Xiaoqin Zhang, Ling Shao, and Shijian Lu.
\newblock A survey of label-efficient deep learning for 3d point clouds.
\newblock \emph{IEEE Transactions on Pattern Analysis and Machine Intelligence}, 2024{\natexlab{a}}.

\bibitem[Xiao et~al.(2024{\natexlab{b}})Xiao, Kai, Zhang, Zha, Sun, and Xiong]{xiao2024event}
Zeyu Xiao, Dachun Kai, Yueyi Zhang, Zheng-Jun Zha, Xiaoyan Sun, and Zhiwei Xiong.
\newblock Event-adapted video super-resolution.
\newblock In \emph{European Conference on Computer Vision}, pages 217--235. Springer, 2024{\natexlab{b}}.

\bibitem[Xie et~al.(2020)Xie, Gu, Guo, Qi, Guibas, and Litany]{xie2020pointcontrast}
Saining Xie, Jiatao Gu, Demi Guo, Charles~R Qi, Leonidas Guibas, and Or Litany.
\newblock Pointcontrast: Unsupervised pre-training for 3d point cloud understanding.
\newblock In \emph{European Conference on Computer Vision}, 2020.

\bibitem[Xin et~al.(2024)Xin, Luo, Zhou, Du, Liu, Fan, Li, and Du]{xin2024parameter}
Yi Xin, Siqi Luo, Haodi Zhou, Junlong Du, Xiaohong Liu, Yue Fan, Qing Li, and Yuntao Du.
\newblock Parameter-efficient fine-tuning for pre-trained vision models: A survey.
\newblock \emph{arXiv preprint arXiv:2402.02242}, 2024.

\bibitem[Xu et~al.(2024)Xu, Kong, Shuai, Zhang, Pan, Chen, Liu, and Liu]{xu2024superflow}
Xiang Xu, Lingdong Kong, Hui Shuai, Wenwei Zhang, Liang Pan, Kai Chen, Ziwei Liu, and Qingshan Liu.
\newblock 4d contrastive superflows are dense 3d representation learners.
\newblock In \emph{European Conference on Computer Vision}, pages 58--80, 2024.

\bibitem[Yang and Wang(2025)]{yang2025kolmogorovarnold}
Xingyi Yang and Xinchao Wang.
\newblock Kolmogorov-arnold transformer.
\newblock In \emph{International Conference on Learning Representations}, 2025.

\bibitem[Yi et~al.(2016)Yi, Kim, Ceylan, Shen, Yan, Su, Lu, Huang, Sheffer, and Guibas]{yi2016scalable}
Li Yi, Vladimir~G Kim, Duygu Ceylan, I-Chao Shen, Mengyan Yan, Hao Su, Cewu Lu, Qixing Huang, Alla Sheffer, and Leonidas Guibas.
\newblock A scalable active framework for region annotation in 3d shape collections.
\newblock \emph{ACM Transactions on Graphics}, 35\penalty0 (6):\penalty0 1--12, 2016.

\bibitem[Yu et~al.(2022)Yu, Tang, Rao, Huang, Zhou, and Lu]{yu2022point}
Xumin Yu, Lulu Tang, Yongming Rao, Tiejun Huang, Jie Zhou, and Jiwen Lu.
\newblock Point-bert: Pre-training 3d point cloud transformers with masked point modeling.
\newblock In \emph{IEEE/CVF Conference on Computer Vision and Pattern Recognition}, 2022.

\bibitem[Zadouri et~al.(2023)Zadouri, {\"U}st{\"u}n, Ahmadian, Ermi{\c{s}}, Locatelli, and Hooker]{zadouri2023pushing}
Ted Zadouri, Ahmet {\"U}st{\"u}n, Arash Ahmadian, Beyza Ermi{\c{s}}, Acyr Locatelli, and Sara Hooker.
\newblock Pushing mixture of experts to the limit: Extremely parameter efficient moe for instruction tuning.
\newblock \emph{arXiv preprint arXiv:2309.05444}, 2023.

\bibitem[Zaken et~al.(2022)Zaken, Goldberg, and Ravfogel]{zaken2022bitfit}
Elad~Ben Zaken, Yoav Goldberg, and Shauli Ravfogel.
\newblock Bitfit: Simple parameter-efficient fine-tuning for transformer-based masked language-models.
\newblock In \emph{Annual Meeting of the Association for Computational Linguistics}, 2022.

\bibitem[Zha et~al.(2023)Zha, Wang, Dai, Chen, Wang, and Xia]{zha2023instance}
Yaohua Zha, Jinpeng Wang, Tao Dai, Bin Chen, Zhi Wang, and Shu-Tao Xia.
\newblock Instance-aware dynamic prompt tuning for pre-trained point cloud models.
\newblock In \emph{IEEE/CVF International Conference on Computer Vision}, 2023.

\bibitem[Zhang et~al.(2023)Zhang, Chen, Bukharin, He, Cheng, Chen, and Zhao]{zhang2023adaptive}
Qingru Zhang, Minshuo Chen, Alexander Bukharin, Pengcheng He, Yu Cheng, Weizhu Chen, and Tuo Zhao.
\newblock Adaptive budget allocation for parameter-efficient fine-tuning.
\newblock In \emph{International Conference on Learning Representations}, 2023.

\bibitem[Zhang et~al.(2022)Zhang, Guo, Gao, Fang, Zhao, Wang, Qiao, and Li]{zhang2022point}
Renrui Zhang, Ziyu Guo, Peng Gao, Rongyao Fang, Bin Zhao, Dong Wang, Yu Qiao, and Hongsheng Li.
\newblock Point-m2ae: multi-scale masked autoencoders for hierarchical point cloud pre-training.
\newblock In \emph{Advances in Neural Information Processing Systems}, 2022.

\bibitem[Zhang et~al.(2024)Zhang, Qi, Dong, Bai, and Wei]{zhang2024positional}
Shaochen Zhang, Zekun Qi, Runpei Dong, Xiuxiu Bai, and Xing Wei.
\newblock Positional prompt tuning for efficient 3d representation learning.
\newblock \emph{arXiv preprint arXiv:2408.11567}, 2024.

\bibitem[Zheng et~al.(2024)Zheng, Huang, Mei, Hou, Lyu, Dai, Ouyang, and Gong]{zheng2024point}
Xiao Zheng, Xiaoshui Huang, Guofeng Mei, Yuenan Hou, Zhaoyang Lyu, Bo Dai, Wanli Ouyang, and Yongshun Gong.
\newblock Point cloud pre-training with diffusion models.
\newblock In \emph{IEEE/CVF Conference on Computer Vision and Pattern Recognition}, pages 22935--22945, 2024.

\bibitem[Zhou et~al.(2024)Zhou, Liang, Xu, Zhu, Xu, Zou, and Bai]{zhou2024dynamic}
Xin Zhou, Dingkang Liang, Wei Xu, Xingkui Zhu, Yihan Xu, Zhikang Zou, and Xiang Bai.
\newblock Dynamic adapter meets prompt tuning: Parameter-efficient transfer learning for point cloud analysis.
\newblock In \emph{IEEE/CVF Conference on Computer Vision and Pattern Recognition}, pages 14707--14717, 2024.

\end{thebibliography}
